\documentclass[preprint,review,3p,authoryear]{elsarticle}

\usepackage{amssymb}

\usepackage{lineno}

\journal{arXiv}

\usepackage{booktabs}
\usepackage{url}
\usepackage[colorlinks]{hyperref}
\usepackage{multirow}
\usepackage{graphicx}
\usepackage{amsmath}
\usepackage{subcaption}
\usepackage[dvipsnames]{xcolor}
\usepackage{bbm}
\usepackage{bm}
\usepackage{soul}
\newcommand{\attack}[1]{\textsf{#1}}

\usepackage{array}
\usepackage{setspace}

\newcolumntype{L}[1]{>{\raggedright\let\newline\\\arraybackslash\hspace{0pt}}m{#1}}
\newcolumntype{C}[1]{>{\centering\let\newline\\\arraybackslash\hspace{0pt}}m{#1}}
\newcolumntype{R}[1]{>{\raggedleft\let\newline\\\arraybackslash\hspace{0pt}}m{#1}}

\newcommand{\hide}[1]{}

\newcommand{\revision}[1]{\textcolor{black}{#1}}
\newcommand{\delete}[1]{\ignorespaces}

\newdefinition{definition}{Definition}

\begin{document}

\begin{frontmatter}



\title{\LARGE{Investigating Imperceptibility of Adversarial Attacks on\\ Tabular Data: An Empirical Analysis}}


\author[inst1,inst2]{Zhipeng He}
\ead{zhipeng.he@hdr.qut.edu.au}

\author[inst1,inst2]{Chun Ouyang}
\ead{c.ouyang@qut.edu.au}
\author[inst2,inst3]{Laith Alzubaidi}
\ead{l.alzubaidi@qut.edu.au}
\author[inst1,inst2]{Alistair Barros}
\ead{alistair.barros@qut.edu.au}
\author[inst4,inst2,inst5]{Catarina Moreira}
\ead{catarina.pintomoreira@uts.edu.au}

\affiliation[inst1]{organization={School of Information Systems, Queensland University of Technology},
            city={Brisbane},
            country={Australia}}

\affiliation[inst2]{organization={Center for Data Science, Queensland University of Technology},
            city={Brisbane},
            country={Australia}}

\affiliation[inst3]{organization={School of Mechanical, Medical and Process Engineering, Queensland University of Technology},
            city={Brisbane},
            country={Australia}}

\affiliation[inst4]{organization={Data Science Institute, University of Technology Sydney},
            city={Sydney},
            country={Australia}}

\affiliation[inst5]{organization={INESC-ID/Instituto Superior Técnico, University of Lisboa},
            city={Lisboa},
            country={Portugal}}

\begin{abstract}









Adversarial attacks are a potential threat to machine learning models by causing incorrect predictions through imperceptible perturbations to the input data. While these attacks have been extensively studied in unstructured data like images, applying them to structured data, such as tabular data, presents new challenges. These challenges arise from the inherent heterogeneity and complex feature interdependencies in tabular data, which differ from the characteristics of image data. To account for this distinction, it is necessary to establish tailored imperceptibility criteria specific to tabular data. However, there is currently a lack of standardised metrics for assessing the imperceptibility of adversarial attacks on tabular data.



To address this gap, we propose a set of key properties and corresponding metrics designed to comprehensively characterise imperceptible adversarial attacks on tabular data. 
These are: 
\textit{proximity} to the original input, 
\textit{sparsity} of altered features, 
\textit{deviation} from the original data distribution, 
\textit{sensitivity} in perturbing features with narrow distribution, 
\textit{immutability} of certain features that should remain unchanged, 
\textit{feasibility} of specific feature values that should not go beyond valid practical ranges, and 
\textit{feature interdependencies} capturing complex relationships between data attributes. 

We evaluate the imperceptibility of five adversarial attacks, including both bounded attacks and unbounded attacks, on tabular data using the proposed imperceptibility metrics. 
The results reveal a \emph{trade-off} between the imperceptibility and effectiveness of these attacks. 
The study also identifies limitations in current attack algorithms, offering insights that can guide future research in the area. 
The findings gained from this empirical analysis provide valuable direction for enhancing the design of adversarial attack algorithms, thereby advancing adversarial machine learning on tabular data. \\ 
\vspace*{-.5\baselineskip}


\end{abstract}






\begin{keyword}
Adversarial examples \sep Imperceptibility \sep Machine learning models \sep Robustness
\end{keyword}

\end{frontmatter}



\newpage
\section{Introduction}
\label{sec:intro}

\emph{Adversarial attacks} aim to generate perturbations on input data, known as \emph{adversarial examples}, to deceive machine learning models into making incorrect predictions~\citep{szegedy2014intriguing}. As a result, adversarial examples can be used to examine the vulnerability and robustness of machine learning models. 
In addition to being effective in causing misclassification, adversarial examples are also expected to contain \emph{imperceptible} perturbations, which can be described as ``indistinguishable to the human eye''~\citep{goodfellow2015explaining}, ensuring they remain undetectable to human observation. 

The imperceptibility of adversarial examples varies significantly between unstructured data (such as images and text) and structured data (like tabular data) due to the inherent differences in their structures and features. Current adversarial attack algorithms primarily target unstructured data, especially images~\citep{croce2020robustbench}. Tabular data distinguishes itself as a prime example of structured data, playing a pivotal role in real-world applications like historical transaction trail in fintech, electronic medical records in healthcare management, and sensor readings in traffic monitoring and control. Compared to image data which is high-dimensional and homogeneous, tabular data is low-dimensional but diverse in feature types and ranges and characterised by heterogeneity and intricate feature interdependencies. 


\begin{figure*}[b!]
    \centering
    \begin{subfigure}[b]{0.49\textwidth}
    \centering
    \includegraphics[width=\textwidth]{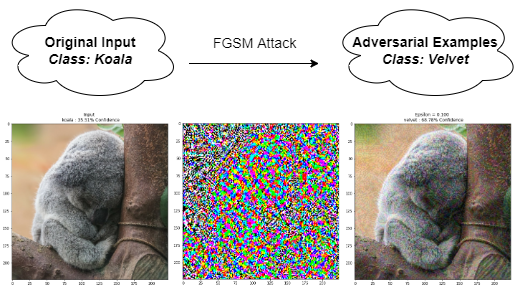}
    \caption{Image Data}
    \label{fig:koala}
    \end{subfigure}
    \begin{subfigure}[b]{0.49\textwidth}
    \centering
    \includegraphics[width=\textwidth]{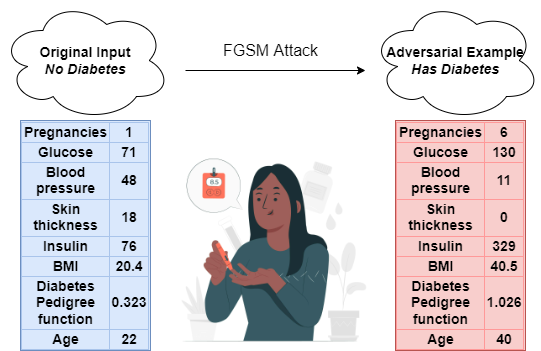}
    \caption{Tabular Data}
    \label{fig:diabetes}
    \end{subfigure}
    \caption{The perturbation on tabular data is more noticeable than images. An adversarial example for a sleeping koala shows how input perturbations caused by a typical adversarial attack, known as Fast Gradient Sign Method (FGSM) attack, can mislead an image recognition system while remaining indistinguishable to human eyes (Figure~\ref{fig:koala}). In contrast, the distinction between two tabular records for classifying the presence of diabetes can easily be observed or detected by humans (Figure~\ref{fig:diabetes}).} 
\end{figure*}

In image data, imperceptibility often refers to making changes to the pixel values that are subtle enough to go unnoticed by the human eye. Adversarial examples are crafted by perturbing pixel values strategically to induce misclassification by machine learning models while maintaining visual similarity to the original~image. 
Consider the example of a specific adversarial attack to an image as depicted in Figure~\ref{fig:koala}: the perturbation applied to the image has resulted in misclassification but is challenging to detect by human observation. However, when applying the same attack algorithm to a tabular dataset as depicted in Figure~\ref{fig:diabetes}, it may effectively cause misclassification, whereas the perturbation-induced change can be easily noticed by humans. 
This suggests that the concept of imperceptibility must be approached differently for tabular data compared to images, as the characteristics of tabular data pose new requirements when addressing imperceptibility. 
These may include: 1) modification of a smaller number of features; 2) minimisation of the distance between the original data points and adversarial examples; and 3) prevention of the generated adversarial examples from being deemed outliers.

Recent developments in adversarial machine learning have increasingly centered on tabular data, with a noticeable emphasis on the imperceptibility of adversarial attacks. Existing research has tended to focus on specific facets of imperceptibility, such as perturbing less significant features determined by feature importance \citep{ballet2019imperceptible}, integrating domain knowledge to safeguard sensitive or immutable features from attacks \citep{Mathov2022not,Chernikova2022FENCE}, managing imperceptibility using cost-constrained attacks \citep{kireev2022adversarial}. A notable gap in the literature lies in the absence of a comprehensive definition for what constitutes imperceptibility specifically in the context of tabular data. 

Therefore, this paper aims to address the gap by providing a more thorough investigation of imperceptibility in adversarial attacks on tabular data. Unlike previous studies that focus on individual aspects of imperceptibility, our research adopts a holistic approach with multifaceted analysis to enhance the understanding of the challenges and opportunities associated with securing tabular data against adversarial attacks. 
Our contributions to the field are twofold. 

Firstly, we propose seven key properties that characterise imperceptible adversarial attacks for tabular data. These properties, namely proximity, sparsity, deviation, sensitivity, immutability, feasibility and feature interdependency, are derived through an examination of the unique characteristics and challenges inherent in tabular data. 
Secondly, we investigate all seven properties of imperceptibility, analyse the relation between adversarial attacks' effectiveness and imperceptibility, and draw insights from a comprehensive evaluation. 
This leads to the findings about limitations of current adversarial attack algorithms in achieving imperceptibility in attacks targeting tabular data. It also helps us identify potential research directions for improving adversarial attack algorithms specifically tailored to tabular data. Furthermore, the insights are essential for both researchers and practitioners, offering tangible guidance in the increasingly complex landscape of adversarial machine learning on tabular data.

The rest of this paper is structured as follows.  
Section~\ref{sec:background} provides an overview of the background and reviews related work on adversarial attacks and their imperceptibility concerning tabular data. 
Section~\ref{sec:metric} proposes imperceptibility properties and metrics pertinent to adversarial attacks on tabular data. 
Section~\ref{sec:evaluation} specifies the experiment settings and analyses the evaluation results. 
Section~\ref{sec:discussion} presents the insights drawn from the evaluation analysis and discusses limitations in current adversarial attack algorithms. Finally, 
Section~\ref{sec:conclusion} concludes the work and outlines potential avenues for future research.




\section{Background and Related Work}
\label{sec:background}

\subsection{Adversarial Attacks and Imperceptibility}

Several state-of-the-art \emph{adversarial attack methods} have been proposed since the concept of adversarial examples was introduced by \citet{szegedy2014intriguing}. 
The \attack{fast gradient sign method (FGSM)} \citep{goodfellow2015explaining} generates adversarial examples by taking the gradient of the model's loss with respect to the input. 
The \attack{basic iterative method (BIM)} \citep{Kurakin2017adversarial} is similar to \attack{FGSM} but aims to find smaller perturbations through iterations. 
The \attack{momentum iterative method (MIM)} and \attack{projected gradient descent (PGD)} are two variants of \attack{BIM}.
\attack{MIM}~\citep{Dong2018boosting} incorporates momentum to better handle noise and increase the transferability of adversarial examples. It keeps a running average of the gradients across iterations, allowing for smoother updates. 
\attack{PGD}~\citep{madry2017towards} iteratively projects the perturbed input onto an $\ell_p$-bounded space around the original input in order to generate more effective attacks targeting at defense mechanisms like gradient masking. 
This makes PGD more robust against gradient masking and stochastic gradient descent attacks. 
The \attack{Jacobian-based saliency map approach (JSMA)} \citep{papernot2016limitation} finds and perturbs the input features that are most important to the model's prediction using saliency map. 
The \attack{Carlini and Wagner (C\&W) Attack}~\citep{carlini2017towards} optimises a loss function that measures the difference between the original inputs and perturbed instances to craft adversarial examples. 
\attack{DeepFool}~\citep{Moosavi2016deepfool} generates adversarial examples by linearising the model's decision boundaries and applying minimal perturbation to move a given input across its nearest boundary. 

\emph{Imperceptibility} is an important property of adversarial perturbations. 
Existing work mainly focus on evaluating imperceptibility of adversarial examples for image data, where $\ell_p$ norm is commonly used to measure the distance between the original and perturbed images~\citep{szegedy2014intriguing}. For example, \citet{Luo2018towards} propose perturbation sensitivity for determining which region of pixels in images is prone to be perceived if being perturbed. Croce and Hein~\citep{Croce2019sparse} use $\ell_0$ norm to limit the number of perturbed pixels in generating sparse and imperceptible adversarial examples. 
Furthermore, \citet{Sharif2018on} suggest considering human perception and semantic information in evaluating the imperceptibility of images, highlighting the limitation of $\ell_p$ norm in measuring imperceptibility from human observation and semantic perspectives. 


\subsection{State-of-the-art Adversarial Attacks on Tabular Data}


\attack{LowProFool}~\citep{ballet2019imperceptible} uses a weighted $\ell_p$ norm to determine the set of features to perturb. This attack method relies on the absolute value of the Pearson's correlation coefficient for each numerical feature to determine their feature importance in order to calculate the weighted $\ell_p$ norm. As such, \attack{LowProFool} is only applicable to numerical features, which restricts its applicability to a specific set of scenarios. \citet{hashemi2020permuteattack} utilise the similarity between adversarial and counterfactual examples, employing counterfactual example generation methods to create adversarial examples. \citet{Mathov2022not} introduce surrogate models that maintain the target model's properties to improve the effectiveness of adversarial example generation techniques for tabular data. 
Alongside traditional distance-based attacks, \citet{kireev2022adversarial} propose a different approach by incorporating the concepts of cost and utility as constraints in the adversarial example generation process. This method relies on a greedy best-first graph search algorithm to effectively craft adversarial examples that meet the relevant cost and utility constraints. 
In addition to white-box attacks, 
\citet{Cartella2021adversarial} adapt three black-box attack techniques, namely the Boundary attack~\citep{brendel2017decision}, HopSkipJump attack~\citep{chen2020hopskipjumpattack}, and ZOO~\citep{chen2017zoo}, to target tabular data.

In the context of tabular data, human eyes tend to more readily detect perturbations in feature values compared to image data. Distance-based metrics alone often cannot fully encompass all the characteristics outlined in Section~\ref{sec:intro}, and additional factors must be considered. For example, the number of features perturbed (or called perturbation size) can also serve as a useful indicator for measuring imperceptibility~\citep{ballet2019imperceptible}. 
Recent studies have also emphasised semantic imperceptibility of adversarial attacks in the context of tabular data~\citep{Mathov2022not,Chernikova2022FENCE}. These works leverage domain knowledge to enhance the efficacy of their algorithms, strategically choosing features to perturb. 
\citet{Mathov2022not} propose rules for immutable features and consider data types for different features. They use an embedding function incorporating all these rules to maintain the validity of perturbed examples. Similarly, \citet{Chernikova2022FENCE} translate domain knowledge related to network traffic into constraints, ensuring that the generated adversarial attacks do not produce invalid adversarial examples. These approaches showcase a subtle integration of domain expertise, shedding light on the intricacies of semantic imperceptibility of adversarial attacks in the context of in tabular data.

Our review indicates that current studies address the imperceptibility of adversarial attacks on tabular data from various angles and with specific methods. However, there is a lack of an overall framework that comprehensively defines imperceptibility in these attacks. This gap highlights the need to not only survey existing approaches but also to establish a unified understanding of imperceptibility in the context of tabular data. This will facilitate a deeper insight into adversarial attacks against tabular data and enable the development of improved adversarial training strategies suitable for such data.



\section{Imperceptibility of Adversarial Attacks on Tabular data}
\label{sec:metric}

%

\subsection{Establishing Criteria for Imperceptibility}

Since the majority of adversarial attack methods have been developed for image data, their application to tabular data requires consideration of the unique characteristics of the latter. In addition to distance-based criteria, it is necessary to account for the specific nature of tabular data, when addressing the imperceptibility of adversarial attacks. Based on the review of literature in Section~\ref{sec:background}, we establish the following criteria in addressing imperceptibility of adversarial attacks on tabular data.

\paragraph{Minimisation of feature perturbation} In principle, smallest changes are expected to help make adversarial examples imperceptible. More specifically, an adversarial example should be crafted as `close' as possible to the original input data points and fewer features should be modified in the perturbation. 

\paragraph{Preservation of statistical data distribution} Adversarial attacks that are expected to be imperceptible should closely align with the input data distribution. This means that the perturbations should maintain the key characteristics of the data on which the model was trained. Adversarial examples that significantly deviate from the original statistical properties are more likely to be detected by the model. 

\paragraph{Narrow-guard feature perturbation}
In tabular data, each feature typically exhibits a unique distribution. When perturbations are applied across features, the impact is more pronounced on features with narrower distributions compared to those with broader distributions, as features with narrower distributions are more sensitive to changes. 
Hence, to generate imperceptible adversarial attacks, it is important to apply perturbations that avoid altering features with narrow distribution or at least minimise their impact. 

\paragraph{Preservation of feature semantics} 
In tabular data, each feature typically has clearly defined semantics and valid practical value ranges, such as gender and age. 
However, adversarial attacks may introduce perturbations that alter feature semantics (e.g., changing gender from female to male) or extend feature values beyond valid practical value ranges (e.g., changing age from 20 to 120). 
Hence, ensuring imperceptibility in adversarial attacks on tabular data requires preserving feature semantics in alignment with its domain knowledge.



\paragraph{Preservation of feature interdependencies} 
Feature interdependency in a tabular dataset refers to the relationships and correlations between different attributes or variables. Recognising feature inter-dependency is important because changes to one feature can affect the validity or interpretation of related features. For example, an individual's age and their age group (such as `young adult' or `senior') are inter-dependent, and changes to the age feature should correspond to changes in the age group feature to maintain consistency. 
Hence, ensuring imperceptibility in adversarial attacks on tabular data requires preserving feature inter-dependencies to maintain data integrity.

\subsection{Properties of Imperceptibility} 

Based on the criteria established above, we propose a set of properties for assessing the imperceptibility of adversarial attacks. 
First of all, we provide a definition of adversarial attacks.


\begin{definition}[Adversarial Attack]
Consider a dataset where each input data point---a vector \(\bm{x} \in \mathbbm{X}\) belongs to a class with label \( y \in \mathbbm{Y}\). 
Let \(f(\cdot)\) denote a machine learning classifier, an adversarial example \(\bm{x}^{adv}\) generated by an adversarial attack is a perturbed input similar to \(\bm{x}\) and misclassifies the label of \(\bm{x}\). 
\begin{equation*}
\bm{x}^{adv} = \bm{x} + \bm{\delta} 
\hspace{0.8em}\text{subject to }  f(\bm{x}^{adv})\neq y
\end{equation*}
\end{definition}
where \(\bm{\delta}\) denotes input perturbation.

\subsubsection{Proximity} 

Given the criterion of \textit{minimisation of feature perturbation}, 
a good adversarial example should introduce \emph{minimal} changes, which can be quantified by maintaining the smallest possible distance from the original feature vector. 
We use the term `proximity' to refer to relevant distance metrics. As is the case with most of the existing adversarial attack algorithms~\citep{carlini2017towards,goodfellow2015explaining,Moosavi2016deepfool}, we employ $\ell_p$ norm to measure the perturbation distance.

\paragraph{Proximity Metrics}
$\ell_p$ norm is a distance metric for measuring the magnitude of a vector in $n$-dimensional space. Three candidatures of $\ell_p$ norm can be used as proximity metrics for measuring the distance between $\bm{x}$ and $\bm{x}^{adv}$ in grid-like path distance ($\ell_1$ distance), straight-line distance ($\ell_2$ distance) and maximum feature difference ($\ell_\infty$ distance).
\begin{equation}
\ell_p(\bm{x}^{adv},\bm{x})=\Vert\bm{x}^{adv}-\bm{x}\Vert_p =\begin{cases}
      \Bigl(\sum_{i=1}^n(x^{adv}_i-x_i)^p \Bigr)^{1/p}, & p\in\{1,2\}\\
      \sup_{n}{\vert x^{adv}_n- x_n\vert}, & p \rightarrow \infty
\end{cases}
\end{equation}

\subsubsection{Sparsity}

Given the criterion of \textit{minimisation of feature perturbation}, 
this property seeks to identify the minimum feature set required to create an adversarial example. Tabular data, unlike images, is relatively low-dimensional, and as such, the number of altered features in a perturbation would have a more obvious effect on its imperceptibility~\citep{Borisov2021deep}. 
Therefore, an ideal adversarial example should change the model's prediction by altering as few features as possible. 

\paragraph{Sparsity Metric}
Straightforwardly, sparsity measures the number of altered features in an adversarial example~$\bm{x}^{adv}$ compared to the original input vector~$\bm{x}$.  
\begin{equation}
    Spa(\bm{x}^{adv}, \bm{x})=\ell_0(\bm{x}^{adv}, \bm{x})=\sum_{i=1}^{n}\mathbbm{1}( x^{adv}_i-x_i)
\end{equation}

\subsubsection{Deviation}
Existing studies~\citep{Lee2018simple,Nguyen2015deep} suggest that adversarial attacks are a special example of \emph{out-of-distribution} being created with the intention to fool a model. Whilst adversarial examples are not deemed to be representative of the actual distribution of predictive models, a perturbed input should be as similar as possible to the majority of original inputs when addressing imperceptibility, thus addressing the criterion of \textit{preservation of statistical data distribution}.  

\paragraph{Deviation Metric} 
We propose to use Mahalanobis distance (MD)~\citep{mclachlan1999mahalanobis} to measure the deviation between an adversarial perturbation and the distribution of the original data input variation. 
MD is a multi-dimensional generalisation of $\ell_2$ norm. This distance metric is commonly used for identifying outliers in multi-dimensional data, as well as for detecting correlations among different features. Given an input vector $\bm{x}$, a perturbed vector $\bm{x}^{adv}$ and the covariance matrix $V$.
\begin{equation}
    \text{MD}(\bm{x}^{adv}, \bm{x})= \sqrt{(\bm{x}^{adv}-\bm{x})V^{-1}(\bm{x}^{adv}-\bm{x})^T}
\end{equation}

\subsubsection{Sensitivity}
\citet{Luo2018towards} propose the notion of perturbation sensitivity for images by calculating the inverse of the standard deviation of the pixel region with the attack perturbation, and add perturbation sensitivity as a regularisation to the attack algorithm to generate imperceptible images. 
In our work, we adapt the concept of perturbation sensitivity as a metric to measure the degree to which the features with narrow distribution in tabular data are altered, addressing the criterion of \textit{narrow-guard feature perturbation}. 


\paragraph{Sensitivity Metric}
Perturbation sensitivity is a normalised $\ell_1$ distance between $\bm{x}$ and $\bm{x}^{adv}$ by the inverse of the standard deviation of all numerical features within the input dataset \(\mathbbm{X}\). It is represented by:
\begin{equation}
\begin{gathered}
       \text{SDV}(x_{i})= \sqrt{ \frac{ \sum^m_j ( x_{i,j}- \bar{x}_{i})^2 }{ m } } \\
           \text{SEN}(\bm{x}, \bm{x}^{adv})=\sum_{i=1}^n \frac{\Vert x^{adv}_{i} - x_{i} \Vert_2}{\text{SDV}(x_i)}
\end{gathered}
\end{equation}
where $n$ is the number of numerical features, $m$ is the number of all input vectors, and $\bar{x}_{i}$ represents the average of the $i\text{th}$ features within all datapoints.


\subsubsection{Immutability}

Given the criterion of \textit{preservation of feature semantics}, when crafting an adversarial example, it should avoid perturbations that could introduce biases or ethical concerns. 
For instance, changing a critical feature like gender, age or ethnic group to influence loan approval would exhibit inherent bias and therefore should not be allowed. Hence, it is important to recognise \textit{immutable features} in tabular data. 

Immutable features are fixed attributes within a dataset that are either inherently unchangeable or should not be altered due to ethical or practical considerations. Examples include personal identifiers, demographic information 
\revision{(such as age or gender)}, and genetic data, which provide a stable foundation for analyses and ensure fairness by preventing biased alterations. 
\revision{For instance, in a medical dataset, a person's date of birth is an immutable feature that should not be manipulated, as altering it could lead to incorrect conclusions or ethical concerns.} 
Preserving the immutability of these features is crucial for maintaining data integrity and ethical compliance. Accordingly, we propose \textit{immutability} as a qualitative property to evaluate whether an attack algorithm can generate adversarial examples that preserve immutable features in tabular data, which contributes to addressing the criterion of \textit{preservation of feature semantics}. 


Some recent works on generating imperceptible adversarial attack for tabular data~\citep{Mathov2022not} attempt to introduce constraints or masks to prevent the modification of immutable features by attack algorithms. 
Extracting these constraints requires a thorough understanding of the relevant domain knowledge and a clear grasp of the specific context of the tabular dataset. However, not all datasets provide the necessary level of detailed context. To this end, using case-based examples can be an effective way to analyse the immutability property in our approach. 


\subsubsection{Feasibility}

Unlike image data, which always have the same fixed value range for pixels (commonly from 0 to 255), tabular data is heterogeneous and often consists of different feature types (e.g., categorical vs. numerical) and varying value ranges. 
When addressing the criterion of \textit{preservation of feature semantics}, adversarial attacks should also prevent the introduction of perturbations that extend feature values beyond feasible practical ranges. 

We employ \textit{feasibility} as a qualitative property to assess whether a perturbed feature value aligns with semantic correctness. It often requires to leverage domain knowledge, if available, or 
\revision{applying} 
common or practical knowledge relevant to the tabular dataset for checking the practical validity of a perturbed feature value. 
\revision{For example, the age of a person should not be perturbed to an unrealistic value, such as 200 years.}
In existing research, feasibility is introduced through the identification of feasible counterfactual examples~\citep{poyiadzi2020face} and adversarial examples~\citep{Chernikova2022FENCE}.

Similar to the analysis of \textit{immutability}, \textit{feasibility} can also be evaluated using case-based examples in our approach. By examining these examples, we can determine if the perturbed feature values maintain their semantic integrity and remain within acceptable practical ranges for a given adversarial attack algorithm. 




\subsubsection{Feature Interdependency} 

Unlike image data, where pixel values are spatially arranged and often interpreted in a relatively straightforward manner by deep learning models, tabular data involves features that may have non-linear and context-specific interactions. Adversarial attacks on tabular data should navigate these intricate dependencies to produce meaningful and realistic perturbations. \revision{For instance, modifying a feature like income in a financial dataset must be consistent with changes in related features such as age and education level to maintain data integrity and realism. However, simply altering income independently of other correlated features could produce anomalies that 
can be easily detected.} 

Hence, we propose \textit{feature interdependency} as a qualitative property 
to measure how well adversarial attack algorithms handle the intricate relationships and dependencies between features within a tabular dataset, thus addressing the criterion of \textit{preservation of feature interdependencies}. 

While feature interdependency may be handled using methods such as feature selection or by imposing constraints that ensure synchronised changes among correlated features, these approaches require careful consideration of the underlying data structure, significant effort in managing data complexity, and a deep understanding of the domain-specific relationships between features. 
Our objective of evaluation is to leverage intrinsic properties of the data to assess whether adversarial attacks incorporate mechanisms to effectively manage interactions between features or attributes in a tabular dataset. 

\section{Evaluation}
\label{sec:evaluation}


\subsection{Design of Experiments}\label{sec:exp-setup}


We conduct a series of experiments to evaluate the imperceptibility of adversarial attack algorithms on tabular datasets, using the properties of imperceptibility proposed in the previous section. 
Our experiment design follows the typical machine learning pipeline, comprising data preparation, model training, and model testing. We generate adversarial examples using adversarial attack algorithms and then assess these algorithms through the following experiments:


\begin{itemize}
    \item Experiment~1: Evaluating the effectiveness of the applied adversarial attack algorithms, specifically how successful these algorithms are in misclassifying a trained machine learning model. 

    \item Experiment~2: Evaluating the imperceptibility of the applied adversarial attack algorithms using quantitative properties, including sparsity, proximity, deviation, and sensitivity.
    \item Experiment 3: Analysing case-based examples to assess the qualitative properties, including immutability, feasibility, and feature interdependency, for the imperceptibility of the applied adversarial attack algorithms.
\end{itemize}

\paragraph{Datasets} We investigate adversarial attacks on five widely-adopted benchmark tabular datasets, namely Adult, German, Breast Cancer, and Diabetes, from UCI library\footnote{\url{https://archive.ics.uci.edu/ml/index.php}}, and COMPAS from ProPublica~\citep{Barenstein2019compas}. Breast Cancer and Diabetes consist of only numerical features, while the other three contain both numerical and categorical features. 

Table~\ref{tab:datasets} presents the main characteristics of five datasets. All datasets are preprocessed to be suitable for binary classification. 
Each dataset is divided into two groups: training set (80\%) used to fit the data to different machine learning models and test set (20\%) to evaluate these models. \delete{and generate adversarial examples.}
\revision{The test set will also be used to generate adversarial examples. For the same dataset, the number of adversarial examples remains consistent across different models and attacks.} Given that the majority of distance-based adversarial attack algorithms employ the $\ell_p$ norm for perturbation size calculation, we utilise a one-hot encoding approach for categorical features to facilitate distance calculation. Ordinal encoding is avoided in our design, as it would introduce an unintended order within categorical features, leading to the potential distortion of distance interpretations.

\begin{table*}[h!]
\caption{Data profiles of five selected tabular datasets.}
\label{tab:datasets}
\centering
\scriptsize{%
\begin{tabular}{@{}ccC{1.3cm}C{1.3cm}C{1.3cm}C{1.2cm}C{1.6cm}C{1.5cm}C{1.5cm}@{}}
\toprule
\textbf{Dataset} & \textbf{Data Type} & \textbf{Total Instances} & \textbf{Train Instances}& \textbf{Test Instances} & \textbf{Total Features} & \textbf{Cate. Features} & \textbf{Num. Features} & \textbf{Encoded Cate. Features }\\ \midrule
Adult Income  & Mixed & 32651  & 26048 & 6513 &  12 & 8 & 4 & 98 \\
Breast Cancer & Num & 569  & 455 & 114   &  30  & 0  & 30  & 0  \\
COMPAS        & Mixed  & 7214  & 5771 & 1443 & 11 & 7 & 4  & 19  \\
Diabetes      & Num  & 768   & 614 & 154   & 8  & 0 & 8  & 0  \\
German Credit & Mixed  & 1000  & 800 & 200   & 20   & 15  & 5  & 58  \\ \bottomrule
\end{tabular}%
}

\end{table*}

\paragraph{Attack Methods} 
To the best of our knowledge, no systematic review or benchmark exists for adversarial attacks on tabular data. To select attack methods for evaluation, we review existing attack benchmarks on image-based attacks and identify those that can be adapted for tabular data. Additionally, most existing attack methods are designed for white-box settings, which means that the attacker has complete knowledge about the target predictive model thus allowing the attacker to craft effective adversarial examples to fool the model. Considering these factors, our selection of attack methods is guided by the following criteria:

\begin{enumerate}
    \item The selected attack methods should be applicable to tabular data.
    \item The selected attack methods should be designed for white-box attacks.
    \revision{\item The selected attack methods should have all essential inputs available in the evaluation.}
    \revision{\item The selected attack methods should be reproducible using open-source code.}
\end{enumerate}

\revision{Due to criteria 3 and 4, We are unable to include two state-of-the-art methods for adversarial attacks on tabular data from our evaluation. \citet{Mathov2022not} does not provide open-source code, making it non-reproducible and incompatible with our requirement for transparency and reproducibility. \citet{Chernikova2022FENCE} requires both datasets and specific constraints as inputs for the algorithm. These constraints rely on detailed background information and domain expertise, which are not available for our datasets sourced from the UCI repository. The absence of such domain knowledge makes it impractical for the evaluation.}

Besides these criteria, we also consider the variety of attack methods, including both \textit{bounded attacks} and \textit{unbounded attacks}. Bounded attacks have an upper bound constraint $\epsilon$ on the \textit{attack budget}, and this constraint limits the magnitude of allowable perturbation that the attack can make to the input data. 
The goal is to find an adversarial example \(\bm{x}^{adv}\), which has perturbation \(\bm{\delta}\) within the budget $\epsilon$ to an input $\bm{x}$, that misleads the prediction of the model being attacked. 
Conversely, unbounded attacks do not consider attack budget, and they attempt to obtain the minimal perturbation $\bm{\delta}$ between input $\bm{x}$ and adversarial example $\bm{x}^{adv}$ that misleads the prediction of the model being attacked. 


We evaluate five adversarial attack algorithms on tabular data, including two bounded attacks (FGSM and PGD) and three unbounded attacks (DeepFool, LowProFool and C\&W Attack). FGSM, PGD, DeepFool and C\&W attacks are primarily developed for image data, whereas their applicability to tabuar data has been under-explored in the existing literature. Nevertheless, theoretical considerations~\citep{goodfellow2015explaining,Moosavi2016deepfool}, including mathematical generalisation of attack mechanisms and their algorithmic adaptability, suggest that these methods can be adapted for perturbing tabular datasets. 
The LowProFool algorithm is specifically tailored for tabular data, with a particular emphasis on numerical features, while neglecting categorical features in its design. As a result, this study will employ LowProFool for the evaluation of two datasets that contain only numerical data. 

To determine the optimal attack configurations, we perform a grid search 
\delete{to find the best hyperparameters and configurations for each attack method.} 
\revision{by systematically exploring a range of hyperparameter values for each attack method.}  
\revision{This involves evaluating different combinations of hyperparameters to identify the settings that yield the best performance for each algorithm.} 
\delete{These parameters aim to optimise the performance and proximity of the respective attack algorithms under investigation.}
\revision{Through this search process, we aim to optimise both the effectiveness and proximity of the attack methods under investigation.}
\revision{The best hyperparameters identified through this process are presented in Table~\ref{tab:hyperparameter}.}


\begin{table}[!htp]
\centering
\caption{\revision{Summary of hyperparameters for bounded attack methods (FGSM, PGD) and unbounded attack methods (DeepFool, LowProFool, C\&W)}}
\label{tab:hyperparameter}
\revision{
\begin{tabular}{@{}cp{4cm}c@{}}
\toprule
\textbf{Attack}                      & \textbf{Hyperparameter}      & \textbf{Value}      \\ \midrule
\multirow{2}{*}{\textit{FGSM}}       & Attack Budget ($\epsilon$)   & 0.3                 \\
                                     & Norm                         & $\ell_\infty$       \\ \midrule
\multirow{3}{*}{\textit{PGD}}        & Attack Budget ($\epsilon$)   & 0.3                 \\
                                     & Norm                         & $\ell_\infty$       \\
                                     & Step Size ($\alpha$)         & 0.1                 \\ \midrule
\multirow{3}{*}{\textit{DeepFool}}   & Norm                         & $\ell_2$            \\
                                     & Max Iterations               & 100                 \\
                                     & Overshoot                    & $1e-6$              \\ \midrule
\multirow{5}{*}{\textit{C\&W}}       & Norm                         & $\ell_2$            \\
                                     & Initial Constant ($c$)       & 0.01                \\
                                     & Max Iterations               & 10                  \\
                                     & Search Steps                 & 10                  \\
                                     & Confidence                   & 0                   \\ \midrule
\multirow{4}{*}{\textit{LowProFool}} & Norm                         & $\ell_2$            \\
                                     & Max Iterations               & 100                 \\
                                     & Trade-off Factor ($\lambda$) & 0.5                 \\
                                     & Feature Importance           & Pearson Correlation \\ \bottomrule
\end{tabular}
}
\end{table}

More specifically, for two bounded attacks (FGSM and PGD), the attack budget~$\epsilon$ is set to 0.3, and the perturbation size is measured using $\ell_\infty$ norm. 
PGD introduces an additional parameter known as the step size~$\alpha$ to control the magnitude of each incremental step during the iterative generation of input perturbation for an attack. 
In our experiments, $\alpha$ is set to 0.1. 

For three unbounded attacks (DeepFool, LowProFool and C\&W attack), the minimal input perturbation is measured, using $\ell_2$ norm, across all three attacks. 
DeepFool is configured with a maximum iteration limit of $100$ and an overshoot parameter of $1e-6$, strategically chosen to strike a balance between maximising the attack's effectiveness and computational costs. 
For the C\&W attack, the initial constant $c$ is set to 0.01 to control the perturbation magnitude, and a maximum of $10$ iterations is performed each involving $10$ search steps to efficiently find the minimal adversarial perturbation. 
The C\&W attack also has a confidence parameter, which influences the confidence level of misclassification, and it is set to 0 in our experiments to avoid any impact of this parameter to the attack effectiveness. 
LowProFool is configured with a maximum iteration limit of 100, and the feature importance is calculated using the Pearson correlation of numerical features (i.e., the default configuration). It also has a trade-off factor~\(\lambda\) in loss function used to balance misclassification and perturbation size, and \(\lambda\) is set to 0.5.


\paragraph{Predictive Models} 
We use three representative machine learning classification models: linear support vector classification (LinearSVC)\footnote{The kernelised Support Vector Classifier (SVC) addresses problems through quadratic programming and does not support loss backpropagation. In contrast, LinearSVC employs hinge loss as its optimisation criterion.}, 
logistic regression (LR) and multi-layer perceptron (MLP), to examine how different models (e.g., linear vs. non-linear models) may affect the performance of adversarial attacks. 
LR is transparent and interpretable, whereas the decision-making process of LinearSVC and MLP is opaque to human. 
It is also worth noting that to identify potential adversarial examples, all chosen white-box attacks require the utilisation of loss gradients to back-propagate the loss function. 
Hence, even though tree-based models like XGBoost have demonstrated strong performance on tabular data~\citep{Shwartz2022tabular}, they are excluded from the experimental design due to their inability to facilitate back-propagation. 

\paragraph{Evaluation Metrics} 
It is meaningful to evaluate the imperceptibility of adversarial attacks only if they demonstrate a reasonable level of effectiveness. Hence, prior to evaluating imperceptibility, it is important to check the effectiveness of adversarial attacks. 
We use established performance metrics, such as \textit{attack success rate},  to assess the effectiveness of adversarial attacks.  
The success rate of an adversarial attack is the ratio of the number of adversarial examples that successfully misclassify a (predictive) model over the total number of adversarial examples~\(m\). 

\begin{equation}
    \text{Success Rate} = \frac{1}{m}\sum_{i=1}^{m}\mathbbm{1}( f(\bm{x}_i^{adv})\neq y_i),
\end{equation}


A higher attack success rate indicates that the assessed attack is more effective. An adversarial attack should achieve a success rate above a specified threshold before its imperceptibility can be evaluated. 
Subsequently, to evaluate the imperceptibility of adversarial attacks, we use quantitative metrics for key properties including proximity, sparsity, deviation and sensitivity. 
We generate multiple adversarial examples for each combination of dataset and model, and then calculate the mean values for these four metrics.
Additionally, the qualitative properties, including immutability, feasibility, and feature interdependency, are assessed through the analysis of case-based examples, offering insights into the practical implications of adversarial attack imperceptibility. 
All the metrics used for the imperceptibility evaluation are detailed in Section~\ref{sec:metric}. 


\paragraph{Implementation} To ensure consistency, all experiments are conducted on a Windows server with an Intel Core i7-11700K CPU and an NVIDIA RTX A4000 GPU. LR and LinearSVC are implemented using \texttt{Sklearn-1.02} 
and NN model is implemented in \texttt{TensorFlow-2.10}. A MLP model consists of five ReLU layers with 24, 12, 12, 12, 12 units respectively, and a softmax layer with two units for outputs. It is trained with 10 epochs and in a batch size of 64. 
All attack algorithms are employed from adversarial attack toolbox \texttt{ART-1.12.2}\footnote{\url{https://github.com/Trusted-AI/adversarial-robustness-toolbox/tree/1.12.2}}. 
The source code is made available in a Github~repository via \url{https://github.com/ZhipengHe/Imperceptibility-of-Tabular-Adversarial-attack}.




\subsection{Evaluation of Attack Effectiveness}


To prepare for evaluating adversarial attacks, the initial phase involves training three distinct predictive models on five datasets. As shown in Table~\ref{tab:accuracy}, subsequent testing confirms that each model achieves satisfactory accuracy, consistently exceeding the 70\% threshold across all datasets. 
Comparing the models within the same dataset, they demonstrate a similar level of performance. These findings validate the feasibility of applying adversarial attacks to these models, and it is viable to assess the effectiveness and imperceptibility of the same adversarial examples across the three predictive models. 

\begin{table}[h!]
\centering
\caption{Model accuracy, precision and recall of three predictive models (LR, LinearSVC and MLP) on five datasets.}
\label{tab:accuracy}
\scriptsize{
\resizebox{\textwidth}{!}{%
\begin{tabular}{@{}l|ccc|ccc|ccc@{}}
\toprule
\multirow{2}{*}{Datasets} & \multicolumn{3}{c}{\textbf{LR}}                          & \multicolumn{3}{c}{\textbf{LinearSVC}}                   & \multicolumn{3}{c}{\textbf{MLP}}                         \\ \cmidrule(l){2-10} 
                          & \textit{Accuracy} & \textit{Precision} & \textit{Recall} & \textit{Accuracy} & \textit{Precision} & \textit{Recall} & \textit{Accuracy} & \textit{Precision} & \textit{Recall} \\ \midrule
Adult                     & 0.8526            & 0.7381             & 0.6028          & 0.8535            & 0.7474             & 0.5933          & 0.8523            & 0.7450             & 0.5894          \\
German                    & 0.8050            & 0.7083             & 0.5763          & 0.8050            & 0.7083             & 0.5763          & 0.7900            & 0.6667             & 0.5763          \\
COMPAS                    & 0.7956            & 0.8166             & 0.9253          & 0.7997            & 0.8148             & 0.9358          & 0.8108            & 0.8275             & 0.9330          \\
Diabetes                  & 0.7662            & 0.7111             & 0.5818          & 0.7662            & 0.6727             & 0.6727          & 0.7532            & 0.6667             & 0.6182          \\
Breast Cancer             & 0.9825            & 1.0000             & 0.9535          & 0.9737            & 0.9762             & 0.9535          & 0.9737            & 0.9545             & 0.9767          \\ \bottomrule
\end{tabular}%
}}
\end{table}

Table~\ref{tab:successrate} presents the results of attack success rates of five different adversarial attack algorithms applied to three predictive models (for classification task) across five datasets. Note that LowProFool is applicable to datasets that contain only numerical data, such as Diabetes and Breast Cancer. Overall, DeepFool is the most effective and C\&W is the least effective. 

\begin{table*}[h!]
\caption{Results of attack success rates of five attack algorithms (FGSM, PGD, C\&W, DeepFool and LowProFool) on three predictive models 
over five datasets. 
The best attack success rate for each model is shown in \textbf{bold}. 
Overall, DeepFool is the most effective, and so is LowProFool but applicable to numerical datasets only. 
FGSM and PGD are generally effective and, in most cases, comparable to DeepFool. 
C\&W is the least effective and performs variably across different models and datasets.}
\label{tab:successrate}
\centering
\scriptsize{%
\begin{tabular}{@{}C{1.7cm}C{2cm}C{2.2cm}C{1.7cm}C{2cm}C{1.7cm}C{2.55cm}@{}}
\toprule
\textbf{Dataset} & \textbf{Model} & \textbf{FGSM} & \textbf{PGD} & \textbf{C\&W} & \textbf{DeepFool} & \textbf{LowProFool} \\ \midrule
\multirow{3}{*}{Adult}      & LR          & \textbf{84.95\%}    & \textbf{84.95\%} & 69.86\%  & 82.69\%           & {--} \\
                            & MLP         & 84.65\%             & \textbf{85.15\%} & 45.19\%  & 84.51\%           & {--} \\
                            & LinearSVC   & 84.75\%             & 84.75\%          & 14.76\%  & \textbf{85.32\%}  & {--} \\ \midrule
\multirow{3}{*}{German}     & LR          & 80.73\%             & 80.73\%          & 72.40\%  & \textbf{81.25\%}  & {--}\\
                            & MLP         & 76.56\%             & 78.65\%          & 61.98\%  & \textbf{80.73\%}  & {--} \\
                            & LinearSVC   & 80.73\%             & 80.73\%          & 18.23\%  & \textbf{81.25\%}  & {--}\\ \midrule
\multirow{3}{*}{COMPAS}     & LR          & 79.26\%             & 79.26\%          & 69.46\%  & \textbf{79.33\%}  & {--} \\
                            & MLP         & 80.82\%             & \textbf{80.89\%} & 72.09\%  & 80.47\%           & {--} \\
                            & LinearSVC   & \textbf{79.76\%}    & \textbf{79.76\%} & 20.81\%  & \textbf{79.76\%}  & {--} \\ \midrule
\multirow{3}{*}{Diabetes}   & LR          & 75.00\%             & 75.00\%          & 75.00\%  & \textbf{78.13\%}  & 75.78\% \\
                            & MLP         & \textbf{72.66\%}    & \textbf{72.66\%} & 71.88\%  & \textbf{72.66\%}  & \textbf{72.66\%} \\
                            & LinearSVC   & \textbf{75.78\%}    & \textbf{75.78\%} & 25.78\%  & \textbf{75.78\%}  & \textbf{75.78\%} \\ \midrule
\multirow{3}{*}{\begin{tabular}[c]{@{}c@{}}Breast \\ Cancer\end{tabular}}
                            & LR          & 96.88\%             & 96.88\%          & 90.63\%  & \textbf{98.44\%}  & \textbf{98.44\%} \\
                            & MLP         & \textbf{96.88\%}    & \textbf{96.88\%} & 82.81\%  & \textbf{96.88\%}  & \textbf{96.88\%}\\
                            & LinearSVC   & \textbf{98.44\%}    & \textbf{98.44\%} & 4.69\%   & \textbf{98.44\%}  & \textbf{98.44\%} \\ \bottomrule
\end{tabular}%
}
\end{table*}

For bounded attacks, FGSM and PGD generally demonstrate competitive performance in attack effectiveness, often achieving high attack success rates comparable to DeepFool. 
For unbounded attacks, DeepFool and LowProFool exhibit consistent performance in attack effectiveness over the Diabetes and Breast Cancer datasets. 
In particular, both achieve near-perfect attack success rates across all models over the Breast Cancer dataset. 
However, it is important to note that LowProFool is only designed for numerical datasets, limiting its applicability to datasets with non-numerical features. 

%
%

In contrast, the C\&W attack yields the lowest attack success rates among all. 
Especially, it demonstrates a consistently low success rate, falling below 30\%, on the LinearSVC model across all datasets. 
One possible reason for this is the optimisation algorithms used in LinearSVC are specifically tailored for the hinge loss function. 
Unlike the other two predictive models LR and MLP, the hinge loss function lacks similar differentiability properties, rendering the C\&W attack ineffective against LinearSVC.

\subsection{Evaluation of Attack Imperceptibility using Quantitative Properties}

As mentioned before, an adversarial attack should achieve a success rate above a reasonable threshold before its imperceptibility can be evaluated. We set a threshold of 30\% for attack success rate. 
As shown in Table~\ref{tab:successrate}, applying the C\&W attack to the LinearSVC model results in a success rate below the specified threshold across all five datasets. 
Therefore, the combination of the C\&W attack on LinearSVC model is excluded from the imperceptibility evaluation. Tables~\ref{tab:sparsity} to~\ref{tab:sensitivity} present the results of imperceptibility evaluation using four quantitative properties.

\paragraph{Sparsity} 

Sparsity metric quantifies the average number of modified features over all adversarial examples generated by an attack. 
A lower sparsity value indicates that an adversarial attack modifies fewer features, suggesting an intention to make the perturbed inputs less noticeable. 
Hence, \emph{lower sparsity suggests better imperceptibility of an adversarial example}. 
For tabular datasets, because the number of features is specific to each dataset, it is meaningful to compare sparsity only within the same dataset, not across different datasets. 
As shown in Table~\ref{tab:sparsity}, the sparsity results indicate that all attack methods yield comparable sparsity cross different models within each dataset. 
In addition, the two bounded attacks, FGSM and PGD, often yield the same or very similar sparsity values. 

\begin{table*}[h!]
\caption{Results of \textbf{\textit{sparsity}} of five attack algorithms (FGSM, PGD, C\&W, DeepFool, and LowProFool) on three predictive models 
over five datasets. 
The best (i.e., lowest) sparsity for each model is shown in \textbf{bold}. 
Note that due to insufficient effectiveness, the combination of C\&W attack on LinearSVC model is not considered and their results are marked with \(\dagger\).}
\label{tab:sparsity}
\centering
\scriptsize{%
\begin{tabular}{@{}C{1.7cm}C{2cm}C{2.2cm}C{1.7cm}C{2cm}C{1.7cm}C{2.55cm}@{}}
\toprule
\textbf{Dataset} & \textbf{Model} & \textbf{FGSM} & \textbf{PGD} & \textbf{C\&W} & \textbf{DeepFool} & \textbf{LowProFool} \\ \midrule
\multirow{3}{*}{Adult}      & LR          & \textbf{3.6779}              & \textbf{3.6779}           & 3.7146   & 3.9497           & {--} \\
                            & MLP         & 3.6890              & 3.6920           & \textbf{3.0948}   & 4.7433           & {--} \\
                            & LinearSVC   & \textbf{3.6946}              & \textbf{3.6946}           & 2.0699\(^\dagger\)   & \textbf{3.6946}            & {--} \\ \midrule
\multirow{3}{*}{German}     & LR          & \textbf{4.3073}              & \textbf{4.3073}           & 5.0104   & 4.6458            & {--}\\
                            & MLP         & 4.4219              & \textbf{4.3281}           & 4.7344   & 6.0000            & {--} \\
                            & LinearSVC   & \textbf{4.3125}              & \textbf{4.3125}           & 3.6927\(^\dagger\)   & \textbf{4.3125}            & {--}\\ \midrule
\multirow{3}{*}{COMPAS}     & LR          & 3.9311              & 3.9311           & \textbf{3.8040}   & 3.9382            & {--} \\
                            & MLP         & 3.9197              & 3.9006            & \textbf{3.8111}   & 3.9929           & {--} \\
                            & LinearSVC   & \textbf{3.9361}              & \textbf{3.9361}           & 3.2003\(^\dagger\)   & \textbf{3.9361}            & {--} \\ \midrule
\multirow{3}{*}{Diabetes}   & LR          & \textbf{7.4375}             & \textbf{7.4375}          & 8.0000  & \textbf{7.4375}             & \textbf{7.4375} \\
                            & MLP         & \textbf{7.5391}              & 7.6250         & 8.0000  & \textbf{7.5391}             & 7.5547 \\
                            & LinearSVC   & \textbf{7.4609}              & \textbf{7.4609}           & 6.8438\(^\dagger\)  & \textbf{7.4609}           & \textbf{7.4609} \\ \midrule
\multirow{3}{*}{\begin{tabular}[c]{@{}c@{}}Breast \\ Cancer\end{tabular}}
                            & LR          & 29.9688             & 29.9688          & \textbf{29.8125}  & 29.9688           & 29.9688 \\
                            & MLP         & 29.8438             & 29.8594           & \textbf{29.8125}  & 29.8438           & 29.9531\\
                            & LinearSVC   & \textbf{29.9375}             & \textbf{29.9375}          & 29.7031\(^\dagger\)   & \textbf{29.9375}         & \textbf{29.9375} \\ \bottomrule
\end{tabular}%
}
\end{table*}

\begin{figure}[!h]
    \begin{subfigure}[b]{0.49\textwidth}
    \centering
    \centering
    \includegraphics[width=\linewidth]{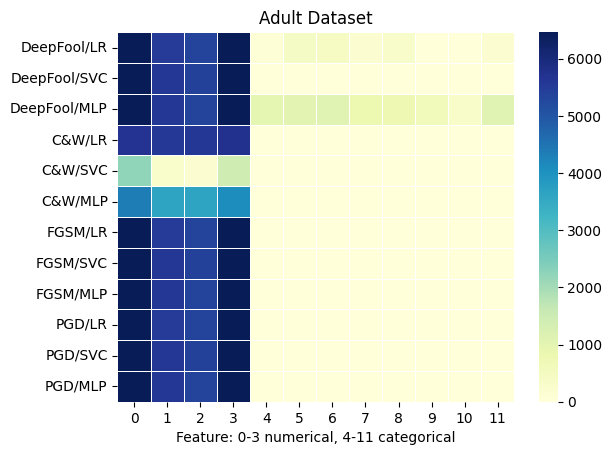}
    \caption{Adult Dataset}
    \label{fig:spa:Adult}
    \end{subfigure}
    \begin{subfigure}[b]{0.49\textwidth}
    \centering
    \centering
    \includegraphics[width=\linewidth]{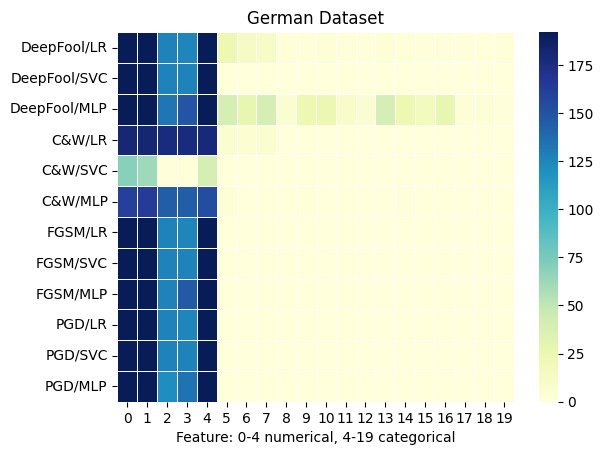}
    \caption{German Dataset}
    \label{fig:spa:German}
    \end{subfigure}
    \begin{subfigure}[t]{\textwidth}
    \centering
    \includegraphics[width=0.5\linewidth]{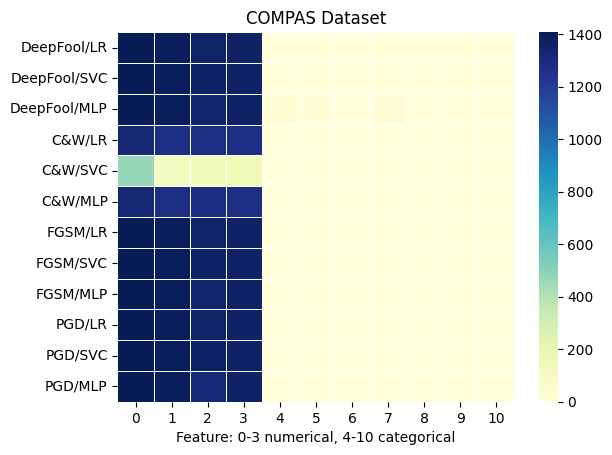}
    \caption{COMPAS Dataset}
    \label{fig:spa:COMPAS}
    \end{subfigure}
    \caption{Three heatmaps visualising the total count of each individual feature being perturbed across different attack/model combinations over three mixed datasets, respectively. The X axis enumerates all the features---numerical features followed by categorical features in each dataset---while the Y axis enumerates all the attack/model combinations. The colour scale of each plot is determined by the number of generated adversarial examples in each individual dataset: 6,513 examples for the Adult dataset, 1,443 examples for the COMPAS dataset, and 200 examples for the German dataset. 
    Overall, the four attacks (DeepFool, C\&W, FGSM and PGD) primarily perturb numerical features with minimal or no changes to categorical features in the three mixed datasets. Due to insufficient effectiveness, the combination of the C\&W attack and LinearSVC is not considered.}
    \label{fig:spa:mixed}
\end{figure}

We also investigate the potential relationship between sparsity and feature types. 
In the two numerical-only datasets (Diabetes and Breast Cancer), we observe that the sparsity value is consistently close to the respective total number of (numerical) features---8 for Diabetes and 30 for Breast Cancer (refer to Table~\ref{tab:datasets}). 
The results suggest that the five attack methods exhibit a tendency to perturb almost all features within these two datasets. 
%
When analysing the three mixed datasets (Adult, German and COMPAS), we first identify which features have been perturbed by comparing the original model inputs and their corresponding adversarial examples for each attack-model combination, and then count the total number of times each feature has been perturbed. 
Due to the varying test sizes of the evaluated datasets, these attacks produce different numbers of adversarial examples across these datasets. 
The results are visualised as heatmaps in Figure~\ref{fig:spa:mixed}, suggesting that the four attacks (DeepFool, C\&W, FGSM and PGD) primaly alter numerical features with minimal or no changes to categorical features in the three mixed datasets. 

\paragraph{Proximity} 
Proximity measures the similarity between the perturbed example and its original input. A smaller proximity metric value indicates that the adversarial example are closer to the original input, suggesting greater imperceptibility. In our evaluation, we use $\ell_2$ and $\ell_\infty$ norm as our two proximity metrics. We refer to them as \textit{proximity $\ell_2$} and \textit{proximity $\ell_\infty$}, and the results of the two metrics are shown in Table~\ref{tab:proximit_l2} (\textit{proximity $\ell_2$}) and Table~\ref{tab:proximit_linf} (\textit{proximity $\ell_\infty$}), respectively. 


From Table~\ref{tab:proximit_l2}, the following observation can be made. 
Overall, the three unbounded attacks, C\&W, DeepFool and LowProFool, yield smaller \textit{proximity \(\ell_2\)} values.
C\&W results in the lowest $\ell_2$ values on both LR and MLP models, with the exception of attacking the LR model on German dataset.
When attacking the LinearSVC model, DeepFool achieves the lowest $\ell_2$ values across all datasets. The results of LowProFool exhibit higher $\ell_2$ values than C\&W and DeepFool. 
The two bounded attacks, FGSM and PGD, have similar $\ell_2$ values on all models within the same dataset. The smaller \(\ell_2\) values of the unbounded attacks suggest their ability to find minimal perturbation without attack budget constraints, while the similarity in \(\ell_2\) values of the bounded attacks (within the same dataset) indicates that the uniform impact of their attack budget constraints is more significant than the influence of different predictive models.

\begin{table*}[t!!!]
\caption{Results of \textbf{\textit{proximity \(\ell_2\)}} of five attack algorithms (FGSM, PGD, C\&W, DeepFool, and LowProFool) on three predictive models 
over five datasets 
The best \textit{proximity \(\ell_2\)} value for each model is shown in \textbf{bold}. 
Due to nsufficient effectiveness, the combination of C\&W attack and LinearSVC is not considered and their results are marked with \(\dagger\).}
\label{tab:proximit_l2}
\centering
\scriptsize{%
\begin{tabular}{@{}C{1.7cm}C{2cm}C{2.2cm}C{1.7cm}C{2cm}C{1.7cm}C{2.55cm}@{}}
\toprule
\textbf{Dataset} & \textbf{Model} & \textbf{FGSM} & \textbf{PGD} & \textbf{C\&W} & \textbf{DeepFool} & \textbf{LowProFool} \\ \midrule
\multirow{3}{*}{Adult}      & LR          & 0.5644              & 0.5644           & \textbf{0.5217}   & 0.6355           & {--} \\
                            & MLP         & 0.5662              & 0.5663           & \textbf{0.1866}   & 0.9627           & {--} \\
                            & LinearSVC   & 0.5656              & 0.5656           & {0.0006}\(^\dagger\)   & \textbf{0.1108}            & {--} \\ \midrule
\multirow{3}{*}{German}     & LR          & \textbf{0.5837}              & \textbf{0.5837}           & 0.6234   & 0.6435            & {--}\\
                            & MLP         & 0.5932              & 0.5826           & \textbf{0.4396}   & 1.1269            & {--} \\
                            & LinearSVC   & 0.5840              & 0.5840           & {0.0001}\(^\dagger\)   & \textbf{0.4128}            & {--}\\ \midrule
\multirow{3}{*}{COMPAS}     & LR          & 0.5289              & 0.5289           & \textbf{0.2426}   & 0.4108            & {--} \\
                            & MLP         & 0.5187              & 0.5109            & \textbf{0.2459}   & 0.4637           & {--} \\
                            & LinearSVC   & 0.5314              & 0.5314           & {0.0017}\(^\dagger\)   & \textbf{0.2360}            & {--} \\ \midrule
\multirow{3}{*}{Diabetes}   & LR          & 0.7842             & 0.7842          & \textbf{0.1942}  & 0.2719             & 0.7545 \\
                            & MLP         & 0.7939              & 0.7999         & \textbf{0.2114}  & 0.2912             & 0.6017 \\
                            & LinearSVC   & 0.7808              & 0.7808           & 0.0091\(^\dagger\)  & \textbf{0.2305}          & 0.6557 \\ \midrule
\multirow{3}{*}{\begin{tabular}[c]{@{}c@{}}Breast \\ Cancer\end{tabular}}
                            & LR          & 1.5263             & 1.5263          & \textbf{0.4392}  & 1.7005           & 1.3016 \\
                            & MLP         & 1.4229             & 1.4672           & \textbf{0.2793}  & 1.5350           & 1.7114\\
                            & LinearSVC   & 1.4978             & 1.4978          & {0.0018}\(^\dagger\)   & \textbf{0.3923}         & 0.9481 \\ \bottomrule
\end{tabular}%
}
\end{table*}
\begin{table*}[b!!!]
\caption{Results of \textbf{\textit{proximity \(\ell_\infty\)}} of five attack algorithms (FGSM, PGD, C\&W, DeepFool, and LowProFool) on three predictive models
over five datasets. 
 The best proximity \(\ell_\infty\) value for each model is shown in \textbf{bold}. Due to insufficient effectiveness, the combination of C\&W attack and LinearSVC is not considered and their results are marked with \(\dagger\).}
\label{tab:proximit_linf}
\centering
\scriptsize{%
\begin{tabular}{@{}C{1.7cm}C{2cm}C{2.2cm}C{1.7cm}C{2cm}C{1.7cm}C{2.55cm}@{}}
\toprule
\textbf{Dataset} & \textbf{Model} & \textbf{FGSM} & \textbf{PGD} & \textbf{C\&W} & \textbf{DeepFool} & \textbf{LowProFool} \\ \midrule
\multirow{3}{*}{Adult}      & LR          & \textbf{0.3000}              & \textbf{0.3000}           & 0.4081   & 0.4189           & {--} \\
                            & MLP         & 0.2999              & 0.3000           & \textbf{0.1570}   & 0.4154           & {--} \\
                            & LinearSVC   & 0.3000              & 0.3000           & {0.0005}\(^\dagger\)   & \textbf{0.1086}            & {--} \\ \midrule
\multirow{3}{*}{German}     & LR          & \textbf{0.3000}              & \textbf{0.3000}           & 0.4146   & 0.3364            & {--}\\
                            & MLP         & \textbf{0.3000}              & \textbf{0.3000}           & 0.3569   & 0.4170            & {--} \\
                            & LinearSVC   & \textbf{0.3000}              & \textbf{0.3000}           & {0.0001}\(^\dagger\)   & 0.3027            & {--}\\ \midrule
\multirow{3}{*}{COMPAS}     & LR          & 0.3000              & 0.3000           & \textbf{0.2147}   & 0.3257            & {--} \\
                            & MLP         & 0.3000              & 0.3000            & \textbf{0.2137}   & 0.3198           & {--} \\
                            & LinearSVC   & 0.3000              & 0.3000           & {0.0015}\(^\dagger\)   & \textbf{0.1833}            & {--} \\ \midrule
\multirow{3}{*}{Diabetes}   & LR          & 0.3000             & 0.3000          & \textbf{0.1491}  & 0.1938             & 0.5228 \\
                            & MLP         & 0.3000              & 0.3000         & \textbf{0.1393}  & 0.1701             & 0.3471 \\
                            & LinearSVC   & 0.3000              & 0.3000           & {0.0066}\(^\dagger\)  & \textbf{0.1592}          & 0.4548 \\ \midrule
\multirow{3}{*}{\begin{tabular}[c]{@{}c@{}}Breast \\ Cancer\end{tabular}}
                            & LR          & 0.3000             & 0.3000          & \textbf{0.1854}  & 0.5240           & 0.4537 \\
                            & MLP         & 0.3000             & 0.3000           & \textbf{0.1502}  & 0.5677           & 0.7110\\
                            & LinearSVC   & 0.3000             & 0.3000          & {0.0009}\(^\dagger\)   & \textbf{0.1408}         & 0.3485 \\ \bottomrule
\end{tabular}%
}
\end{table*}

Table~\ref{tab:proximit_linf} shows that the \textit{proximity $\ell_\infty$} values for both FGSM and PGD attacks are consistently 0.3 across all models and datasets. This is expected since the $\ell_\infty$ values for bounded attacks are constrained by the attack budget settings outlined in Section \ref{sec:exp-setup}. 
For C\&W and DeepFool, the lowest $\ell_\infty$ values are achieved in most cases, except for the German dataset. Similar to the observation on \textit{proximity \(\ell_2\)}, the C\&W attack yields the lowest $\ell_\infty$ values on both LR and MLP models, while DeepFool consistently outperforms on the LinearSVC model. In contrast, LowProFool produces the highest $\ell_\infty$ values among unbounded attacks on the Diabetes and Breast Cancer datasets. 



\paragraph{Deviation}
Deviation metric assesses how perturbed examples relate to the overall dataset distribution. When the deviation metric is notably high, adversarial attacks are more likely to produce examples outside of the dataset's distribution. Therefore, lower deviation values are preferable for crafting imperceptible attacks.
In Table~\ref{tab:deviation}, the evaluation of the deviation metric reveals that C\&W consistently yields the lowest deviation value for both LR and MLP models, implying a higher degree of similarity to the corresponding input dataset's statistical distribution. When attacking the LinearSVC model, DeepFool produces the lowest deviation value. 
The deviation value of LowProFool is higher than that of C\&W and DeepFool, but lower than the two bounded attacks. 
In contrast, the two bounded attacks, FGSM and PGD, share the similar deviation values and consistently yield higher deviation values, suggesting that they tend to produce more out-of-distribution data. 

\begin{table*}[h!!!]
\caption{Results of \textbf{\textit{deviation}} of five attack algorithms (FGSM, PGD, C\&W, DeepFool, and LowProFool) on three predictive models 
over five datasets. 
The best deviation for each model is shown in \textbf{bold}. Due to insufficient effectiveness, the combination of C\&W attack and LinearSVC is not considered and their results are marked with \(\dagger\).}
\label{tab:deviation}
\centering
\scriptsize{%
\begin{tabular}{@{}C{1.7cm}C{2cm}C{2.2cm}C{1.7cm}C{2cm}C{1.7cm}C{2.55cm}@{}}
\toprule
\textbf{Dataset} & \textbf{Model} & \textbf{FGSM} & \textbf{PGD} & \textbf{C\&W} & \textbf{DeepFool} & \textbf{LowProFool} \\ \midrule
\multirow{3}{*}{Adult}      & LR          & 0.0787              & 0.0787           & \textbf{0.0773}   & 0.1274           & {--} \\
                            & MLP         & 0.0779              & 0.0786           & \textbf{0.0277}   & 0.2861           & {--} \\
                            & LinearSVC   & 0.0788              & 0.0788           & {0.0001}\(^\dagger\)   & \textbf{0.0089}            & {--} \\ \midrule
\multirow{3}{*}{German}     & LR          & \textbf{0.1519}              & \textbf{0.1519}           & 0.1540   & 0.2109            & {--}\\
                            & MLP         & 0.1654              & 0.1595           & \textbf{0.0999}   & 0.4722            & {--} \\
                            & LinearSVC   & 0.1521              & 0.1521           & {0.0000}\(^\dagger\)   & \textbf{0.0859}            & {--}\\ \midrule
\multirow{3}{*}{COMPAS}     & LR          & 0.0576              & 0.0576           & \textbf{0.0270}   & 0.0513            & {--} \\
                            & MLP         & 0.0568              & 0.0563            & \textbf{0.0277}   & 0.0856           & {--} \\
                            & LinearSVC   & 0.0577              & 0.0577           & {0.0002}\(^\dagger\)   & \textbf{0.0290}            & {--} \\ \midrule
\multirow{3}{*}{Diabetes}   & LR          & 0.1375             & 0.1375          & \textbf{0.0326}  & 0.0500             & 0.1365 \\
                            & MLP         & 0.1463              & 0.1492         & \textbf{0.0344}  & 0.0560             & 0.1128 \\
                            & LinearSVC   & 0.1342              & 0.1342           & {0.0014}\(^\dagger\)  & \textbf{0.0384}          & 0.1081 \\ \midrule
\multirow{3}{*}{\begin{tabular}[c]{@{}c@{}}Breast \\ Cancer\end{tabular}}
                            & LR          & 0.7075             & 0.7075          & \textbf{0.2088}  & 0.8427           & 0.6694 \\
                            & MLP         & 0.2807             & 0.3552           & \textbf{0.0623}  & 0.4171           & 0.4691\\
                            & LinearSVC   & 0.5639             & 0.5639          & {0.0006}\(^\dagger\)   & \textbf{0.1540}         & 0.3788 \\ \bottomrule
\end{tabular}%
}
\end{table*}

\paragraph{Sensitivity}
Sensitivity quantifies how much an attack algorithm perturbs those features with low variability (i.e. features with a narrow distribution). 
Lower sensitivity values are preferable because attack algorithms would introduce fewer perturbation on features with a narrow distribution, which in turn reduces imperceptibility.
As shown in Table~\ref{tab:sensitivity}, for each individual dataset, two bounded attacks generally exhibit similar values in deviation, which are higher than those observed for C\&W and DeepFool attacks. However, LowProFool, an unbounded attack, shows similar results to FGSM and PGD.
Regarding LR and MLP models, the C\&W attack demonstrates lowest values in sensitivity metric with exception of the combination of German dataset and LR model. 
DeepFool always yeilds the lowest sensitivity values in all five dataset on LinearSVC. 
This indicates that LowProFool, FGSM, and PGD attacks tend to introduce more significant perturbations on narrow distribution features than C\&W and DeepFool.

\begin{table*}[]
\caption{Results of \textbf{\textit{sensitivity}} of five attack algorithms (FGSM, PGD, C\&W, DeepFool, and LowProFool) on three predictive models 
over five datasets. 
The best sensitivity for each model is shown in \textbf{bold}. Due to insufficient effectiveness, the combination of C\&W attack and LinearSVC is not considered and their results are marked with \(\dagger\).}
\label{tab:sensitivity}
\centering
\scriptsize{%
\begin{tabular}{@{}C{1.7cm}C{2cm}C{2.2cm}C{1.7cm}C{2cm}C{1.7cm}C{2.55cm}@{}}
\toprule
\textbf{Dataset} & \textbf{Model} & \textbf{FGSM} & \textbf{PGD} & \textbf{C\&W} & \textbf{DeepFool} & \textbf{LowProFool} \\ \midrule
\multirow{3}{*}{Adult}      & LR          & 2.4898              & 2.4898           & \textbf{1.5780}   & 2.3923           & {--} \\
                            & MLP         & 2.4998              & 2.5019           & \textbf{0.4857}   & 2.6746           & {--} \\
                            & LinearSVC   & 2.5020              & 2.5020           & {0.0016}\(^\dagger\)   & \textbf{0.4252}            & {--} \\ \midrule
\multirow{3}{*}{German}     & LR          & 1.1172              & 1.1172           & 1.0668   & \textbf{0.9765}            & {--}\\
                            & MLP         & 1.1397              & 1.1130           & \textbf{0.6896}   & 1.0908            & {--} \\
                            & LinearSVC   & 1.1178              & 1.1178           & {0.0002}\(^\dagger\)   & \textbf{0.8187}            & {--}\\ \midrule
\multirow{3}{*}{COMPAS}     & LR          & 2.8215              & 2.8215           & \textbf{0.7602}   & 1.6436            & {--} \\
                            & MLP         & 2.6871              & 2.5944            & \textbf{0.7803}   & 1.8902           & {--} \\
                            & LinearSVC   & 2.8485              & 2.8485           & {0.0047}\(^\dagger\)   & \textbf{0.9090}            & {--} \\ \midrule
\multirow{3}{*}{Diabetes}   & LR          & 1.6910             & 1.6910          & \textbf{0.2931}  & 0.4780             & 1.3519 \\
                            & MLP         & 1.7464              & 1.7691         & \textbf{0.3613}  & 0.5704             & 1.1873 \\
                            & LinearSVC   & 1.6867              & 1.6867           & {0.0147}\(^\dagger\)  & \textbf{0.4114}          & 1.1751 \\ \midrule
\multirow{3}{*}{\begin{tabular}[c]{@{}c@{}}Breast \\ Cancer\end{tabular}}
                            & LR          & 1.9884             & 1.9884          & \textbf{0.3773}  & 1.8599           & 1.3337 \\
                            & MLP         & 1.8176             & 1.9000           & \textbf{0.2367}  & 1.6574           & 1.8253\\
                            & LinearSVC   & 1.9371             & 1.9371           & {0.0016}\(^\dagger\)  & \textbf{0.4496}           & 1.0620\\ \bottomrule
\end{tabular}%
}
\end{table*}

\paragraph{Summary}

The evaluation of the four quantitative imperceptibility properties suggests that overall the three unbounded attacks, C\&W, DeepFool and LowProFool, generate more imperceptible adversarial examples compared to the two bounded attacks, FGSM and PGD. 
C\&W often achieves the best imperceptibility on LR and MLP across different datasets, and DeepFool yields the best imperceptibility on LinearSVC. 
In contrast, LowProFool demonstrates the worst imperceptibility among three unbounded attacks in our evaluation. 
The two bounded attacks, FGSM and PGD, exhibit similar results across each property evaluation, likely due to their shared attack mechanism, differing only in their one-time versus iterative application.

\subsection{Analysis of Imperceptibility using Qualitative Properties}

In Section~\ref{sec:metric}, we define three qualitative properties of attack imperceptibility, which are \textit{Immutability}, \textit{feasibility} and \textit{feature interdependency}, and they often rely on domain-specific knowledge about the datasets.
\delete{We analyse these qualitative properties using case-based examples.}
\revision{The systematic process for analysing and evaluating adversarial examples based on these qualitative properties involves the following key points:}

\paragraph{\revision{\textbf{Step 1:} Analyse and Identify Key Features}}
\begin{itemize}
    \item \revision{\textit{Immutability:} Identify features that must remain unchanged during any modification. This requires a thorough understanding of domain-specific knowledge. For example, certain attributes like gender and age must remain unchanged in the context of the dataset. }
    \item \revision{\textit{Feasibility:} Define the feasible range for each modifiable feature, ensuring the values stay realistic within the data domain. This can include upper and lower bounds, categorical constraints, or logical limitations specific to the dataset. While statistical methods can be employed to identify frequent and infrequent values, determining whether infrequent values are truly infeasible requires relevant domain knowledge.}
    \item \revision{\textit{Feature Interdependency:} Analyse the relationships or correlations between different features, ensuring modifications respect these dependencies. For instance, an increase in one feature may necessitate a corresponding increase or decrease in another, depending on the domain-specific rules. While statistical methods can reveal correlations between individual features, domain knowledge is essential to identify and understand more complex interdependencies.}
\end{itemize}
\paragraph{\revision{\textbf{Step 2:} Evaluate Adversarial Examples Against Properties}}
\begin{itemize}
    \item \revision{\textit{Immutability}: Does the attack alter features that should remain unchanged?}
    \item \revision{\textit{Feasibility}: Are the modifications within the feasible ranges of the corresponding features?}
    \item \revision{\textit{Feature Interdependency}: Are the feature interdependencies between features preserved after the attack?}
\end{itemize}

\revision{Our current evaluation prioritises the use of domain knowledge to assess the qualitative properties of immutability, feasibility, and feature interdependency. Statistical techniques, such as identifying common value ranges and detecting correlations between features, can offer an additional layer of objectivity and precision. These methods will be explored and incorporated in future work.}

\revision{In our current work, for immutability, we applied common knowledge as domain knowledge and have checked all the adversarial examples generated by each of the evaluated attacks on all three datasets containing mixed variables (i.e. Adult, COMPAS, and German). For feasibility, we used the collected domain knowledge (see Table~\ref{tab:diabetes-knowledge}) to identify the feasible value ranges for each feature in the Diabetes dataset and evaluated all the adversarial attacks to check if any perturbed features fall outside these ranges. We also observed apparent feature interdependencies in the COMPAS dataset and evaluated whether the generated adversarial examples violate these interdependencies.}

\revision{When presenting the evaluation results, it is impractical to explain the results for all three properties across every combination of dataset, model, and attack. Therefore, we have selected a few representative examples to illustrate the key findings on immutability, feasibility, and feature interdependency.}

\paragraph{Immutability}\label{sec:immutable}


We analyse immutability across three mixed datasets---Adult, COMPAS and German, which we can identify examples of immutable features using common knowledge. These examples of immutable features in each of these datasets are shown below: 
\begin{itemize}
    \item Adult: \textit{Sex}, \textit{Race} and \textit{Marital Status}
    \item COMPAS: \textit{Sex} and \textit{Race}
    \item German: \textit{Personal Status \& Sex} and \textit{Foreign Worker Status}
\end{itemize}

\begin{figure}
    \centering
    \includegraphics[width=\linewidth]{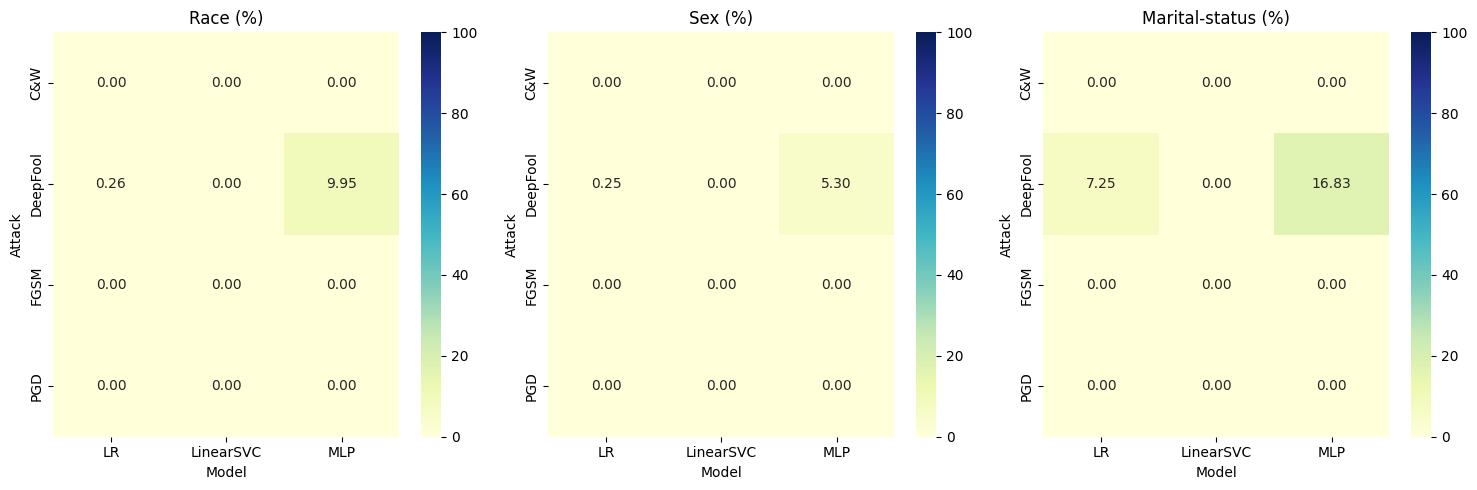}
    \caption{\revision{Heatmap showing the percentage of adversarial examples generated for each combination of model (LR, LinearSVC, MLP) and attack type (DeepFool, C\&W, FGSM, PGD) that perturb immutable features: race, sex, and marital-status. The values are presented as a percentage of a total of 6513 adversarial examples.}}
    \label{fig:immutable-adult}
\end{figure}

We examine each adversarial example to check whether the above immutable feature values have been modified. This allows us to assess whether the identified immutable features are susceptible to manipulation. The results of perturbing immutable features for each of the three datasets are detailed in \ref{sec:appendix:immutable}. 
%
We observe that DeepFool successfully perturbs immutable features in both LR and MLP models across all three datasets, violating the immutability requirement. 
For instance, in the Adult dataset \revision{(Figure~\ref{fig:immutable-adult})}, DeepFool generates a total of 6,513 adversarial examples when attacking the MLP model. In these examples, the feature \textit{Race} is modified in 648 instances \revision{(9.95\%)}, \textit{Sex} in 345 instances \revision{(5.3\%)}, and \textit{Marital Status} in 1,096 instances \revision{(16.83\%)}.

Conversely, C\&W, FGSM and PGD do not appear to alter these immutable features. However, this does not necessarily mean that the three attacks comply with the immutability requirement. 
Theoretically, the design of these attacks does not consider any strategy for avoiding perturbation of immutable features. 
A close examination of these attacks has revealed that they are incapable of changing immutable features---which are all categorical features in the examples used and are preprocessed with one-hot encoding in implementation.


To investigate how immutable features are perturbed by DeepFool, we use the COMPAS dataset and LR model as an example to illustrate how DeepFool perturbs immutable features. The COMPAS dataset, known for its racial bias in predicting the likelihood of recidivism either as ``Medium-Low'' or ``High'' risk~\citep{Barenstein2019compas}. In our experiment, \textit{race} and \textit{sex} are considered as immutable features. We generate 1,408 adversarial examples for DeepFool on the LR model, and retrieve 8 cases with perturbation on \textit{race} (0 case for \textit{sex}).
Table~\ref{tab:immutable} provides two example cases (\#285 and \#501) out of these 8 cases, which are correctly predicted as ``Medium-Low'' by the LR model. DeepFool successfully misclassified the predictions of these two cases by altering the value of the immutable feature \textit{race} from ``Caucasian'' to ``Native American'' (Case \#285) and from ``Hispanic'' to ``African-American'' (Case \#501), respectively. 

\begin{table*}[h!]
\caption{Partial feature values between original datapoints and adversarial examples generated by DeepFool attack on LR in Case \#285 and Case \#501 over the COMPAS dataset.}
\label{tab:immutable}
\centering
\scriptsize{%
\begin{tabular}{@{}ccC{0.7cm}C{1cm}C{1cm}cC{1cm}cc@{}}
\toprule
\textbf{Case}                   & \textbf{Attack}   & \textbf{Age} & \textbf{Priors Count} & \textbf{Length of Stay} & \textbf{Age Cat.} & \textbf{Sex}  & \textbf{Race}             & \textbf{Class} \\ \midrule
\multirow{2}{*}{\#285} & Original & 80  & 0                                                      & 0                                                        & Greater than 45 & Male & \textit{Caucasian}        & \textit{Medium-Low}\\ \cmidrule(l){2-9} 
                       & DeepFool & 18  & 38                                                     & 799                                                      & Greater than 45 & Male & \textit{Native American}  & \textit{High}\\ \midrule
\multirow{2}{*}{\#501} & Original & 83  & 0                                                      & 0                                                        & Greater than 45 & Male & \textit{Hispanic}         & \textit{Medium-Low}\\ \cmidrule(l){2-9} 
                       & DeepFool & 18  & 38                                                     & 799                                                      & Less than 25    & Male & \textit{African-American} & \textit{High}\\ \bottomrule
\end{tabular}%
}
\end{table*}

\begin{figure}[h!]
    \centering
    \centering
    \includegraphics[width=0.5\textwidth]{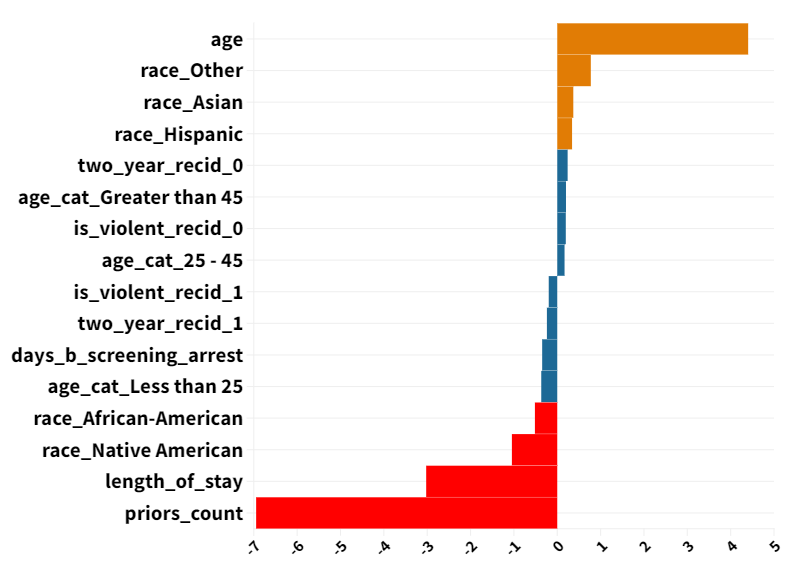}
    \caption{Feature weights (coefficients) of LR for the COMPAS dataset. The features are ranked by feature weights. Features with a positive weight contribute to ``Medium-Low'' risk and features with a negative weight contribute to ``High'' risk. Features with a weight in the range of $[-0.15,0.15]$ have little impact on model prediction and therefore are omitted.}
    \label{fig:immutable}
\end{figure}

We also examine the feature weights (coefficients) of the LR model, as depicted in Figure~\ref{fig:immutable}. 
Considering the two example cases in Table~\ref{tab:immutable}, DeepFool tends to perturb features with considerable impacts on model prediction, and \textit{race} is among these features. In contrast, the other immutable feature, \textit{sex}, has little impact on model prediction, and therefore it is not modified by DeepFool.

\paragraph{Feasibility}

This property assesses if adversarial attack algorithms can perturb feature values within a reasonable range. However, since there is a shortage of background knowledge for all datasets, it is impractical to identify all feasible feature ranges for each dataset. To assess the feasibility of adversarial attacks, we turn to the Diabetes dataset. It only comprises numerical features capturing diagnostic measurements which are used to predict diabetes. 
According to the description of each variable in the Diabetes dataset~\citep{smith1988using},
we identify feasible value ranges for all features and these are listed in \ref{sec:appendix:feasible}.

\begin{table*}[b!]
\caption{Partial feature values between original datapoints and adversarial examples generated by DeepFool, C\&W and LowProFool attacks in Case \#19 and Case \#57 over the Diabetes dataset.}
\label{tab:feasibility}
\centering
\scriptsize{%
\begin{tabular}{@{}C{0.7cm}cC{1.3cm}C{1.3cm}C{1.5cm}C{1cm}C{1cm}C{1cm}C{1.4cm}@{}}
\toprule
\textbf{Case}                  & \textbf{Attack}        & \textbf{Glucose} & \textbf{Blood Pressure} & \textbf{Skin Thickness} & \textbf{Insulin} & \textbf{BMI} & \textbf{Age} & \textbf{Diabetes?}\\ \midrule
\multirow{4}{*}{\#19} & Original      & 86.00                       & 68.00                                                                        & 28.00                                                                        & 71.00                       & 30.20                   & 24.00                   & N\\ \cmidrule(l){2-9} 
                      & DeepFool      & 159.50                      & 62.21                                                                        & 30.24                                                                        & 51.32                       & 49.00                   & 32.69                   & Y\\ \cmidrule(l){2-9} 
                      & C\&W       & 135.98                      & 63.93                                                                        & 29.18                                                                        & 69.29                       & 43.22                   & 24.32                  & Y \\ \cmidrule(l){2-9} 
                      & LowProFool & 199.00                      & 56.17                                                                        & 32.57                                                                        & 30.80                       & 67.10                   & 41.74                  & Y \\ \midrule
\multirow{4}{*}{\#57} & Original      & 74.00                       & 68.00                                                                        & 28.00                                                                        & 45.00                       & 29.70                   & 23.00                  & N \\ \cmidrule(l){2-9} 
                      & DeepFool      & 191.86                      & 58.71                                                                        & 31.59                                                                        & 13.43                       & 59.85                   & 36.93                  & Y \\ \cmidrule(l){2-9} 
                      & C\&W       & 136.88                      & 62.84                                                                        & 29.65                                                                        & 43.81                       & 45.97                   & 23.24                  & Y \\ \cmidrule(l){2-9} 
                      & LowProFool & 199.00                      & 55.53                                                                        & 32.81                                                                        & 2.65                        & 67.10                   & 41.69                  & Y \\ \bottomrule
\end{tabular}%
}

\end{table*}

To investigate how adversarial attacks modify features beyond their feasible ranges, we select two case examples \#19 and \#57 from the generated adversarial examples with the LR model. We then compare the original feature values with the adversarial examples generated by DeepFool, C\&W, and LowProFool attack in Table~\ref{tab:feasibility}. The following observations can be made: 
\begin{itemize}
    \item \textit{Glucose}: In both cases, all three attack methods increase the value of \textit{Glucose} from below 90 to above 135. The C\&W attack introduces comparatively minor perturbations, and the revised \textit{Glucose} values are still within the normal range (below 140). 
    The DeepFool and LowProFool attacks introduce more substantial changes, pushing the glucose levels into the impaired glucose tolerance range (140--199). Particularly with LowProFool, it alters the \textit{Glucose} value to reach 199 in both cases, which is extremely close to the glucose threshold (200) indicating diabetes.
    \item \textit{BloodPressure}: None of the adversarial examples surpass the normal threshold (80) for \textit{BloodPressure}.
    \item \textit{SkinThickness}: The adversarial examples generated by DeepFool in Case \#57 and by LowProFool in both cases exceed the maximum reference value (23.6$\pm$7.5mm). 
    \item \textit{Insulin}: In Case \#57, both the DeepFool and LowProFool attacks decrease the insulin level to below the lower bound of the recommended 2-hour serum insulin range (16--166). This perturbed feature value is not feasible regarding the expected physiological behavior.
    \item \textit{BMI}: In both cases, all attacks modify the \textit{BMI} values from around 30 to over 40, indicating severe obesity. Particularly in Case \#57, the DeepFool and LowProFool attacks perturb the \textit{BMI} to 59.85 and 67.10, respectively, which are unrealistic.
\end{itemize}


\begin{figure}[h!]
    \centering
    \includegraphics[width=0.5\textwidth]{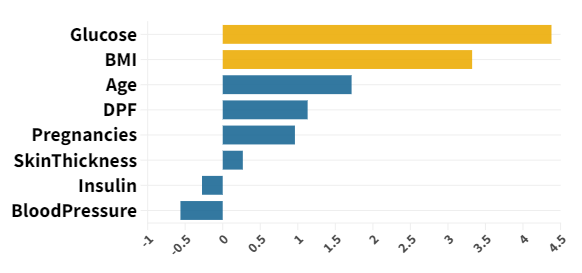}
    \caption{Feature weights (coefficients) of LR for the Diabetes dataset. The features are ranked by feature weights. Features with a positive weight contribute to ``has'' diabetes and features with a negative weight contribute to ``not have'' diabetes. }
    \label{fig:feasibility}
\end{figure}

The feature weights of the LR model for the Diabetes dataset are visualised in Figure~\ref{fig:feasibility}. It is evident that \textit{Glucose} and \textit{BMI} are the top two most important features contributing to the prediction of having diabetes. This explains why all the above three attacks tend to introduce significant perturbations in these two features, often pushing their resulting feature values into infeasible ranges.


\paragraph{Feature Interdependency}

Among the five mixed datasets, we consider the COMPAS dataset for analysis since the dataset exhibits apparent interdependencies among features. 
Specifically, the features \textit{Age} and \textit{Age Category} capture similar information but serve different purposes: 
\textit{Age} represents an individual's exact age, while \textit{Age Category} groups individuals into broader age ranges for analytical purposes. 
Maintaining consistency between these two features is essential for preserving the dataset's integrity. 
Therefore, if the value of \textit{Age} is altered, the \textit{Age Category} should be correspondingly updated to reflect this change accurately.

Revisiting the two adversarial examples of the DeepFool attack in Table~\ref{tab:immutable}, we observe the following issue in Case \#285. When \textit{Age} was changed from 80 to 18, the \textit{Age Category} was not updated accordingly and remained at ``Greater than 45'', breaking the interdependency between these two features. This discrepancy arises because DeepFool does not account for feature interdependencies when perturbing input data. Therefore, although in Case \#501 the \textit{Age Category} was adjusted consistently with the modified \textit{Age}, this alignment is most likely coincidental.

\section{Discussion} 
\label{sec:discussion}

\subsection{Imperceptibility of Bounded vs. Unbounded Attacks}


Despite their high attack success rates, the two bounded attacks---FGSM and PGD---often struggle to maintain good imperceptibility, especially when considering metrics such as deviation and sensitivity. 
While a pre-defined attack budget is a key constraint affecting attack performance, determining an appropriate budget for bounded attacks on tabular data presents a significant challenge. In theory, systematically exploring different attack budget values and evaluating their impact on both success rate and imperceptibility metrics could reveal the optimal balance that maximises attack effectiveness while minimising perceptibility. However, conducting an exhaustive search to determine the best attack budget for each dataset is impractical due to resource limitations and the vast diversity of datasets. 


In contrast, unbounded attacks are more likely to craft imperceptible adversarial examples. C\&W and DeepFool outperform FGSM and PGD in the evaluation of proximity, deviation and sensitivity. Nevertheless, further improvements are necessary for the C\&W attack to enhance their performance in attack success rate of attacking tabular dataset.

Moreover, none of these algorithms systematically address the requirements of feasibility, immutability, and feature interdependency in tabular data. While the design of these algorithms does not inherently consider these factors, auxiliary methods, such as feature masks, can be used during the implementation of the attack algorithms to preserve immutable features or control features in feasible ranges.
One potential improvement is to integrate these qualitative imperceptibility constraints directly into the algorithm's design.

\subsection{Adversarial Attacks' Imperceptibility vs Effectiveness}

According to our evaluation on the effectiveness and imperceptibility of five adversarial attack algorithms, none of them can achieve both the highest attack success rate and the best imperceptibility at the same time. 
In order to investigate the relationship between the imperceptibility and effectiveness of adversarial attacks on tabular data, we analyse the correlation between the attack success rate and four quantitative properties related to imperceptibility for all attacks. In general, all five adversarial attacks share similar sparsity values within the same dataset, attempting to perturb all numerical features in the dataset regardless of attacks' effectiveness. 
For the other three imperceptibility properties---proximity, deviation and sensitivity---we categorise the adversarial examples into two groups: \emph{successful} and \emph{unsuccessful} examples. Box plots are then used to compare the distribution of individual metrics across the two groups of adversarial examples. 

\begin{figure*}[]
    \centering
    \includegraphics[width=\textwidth]{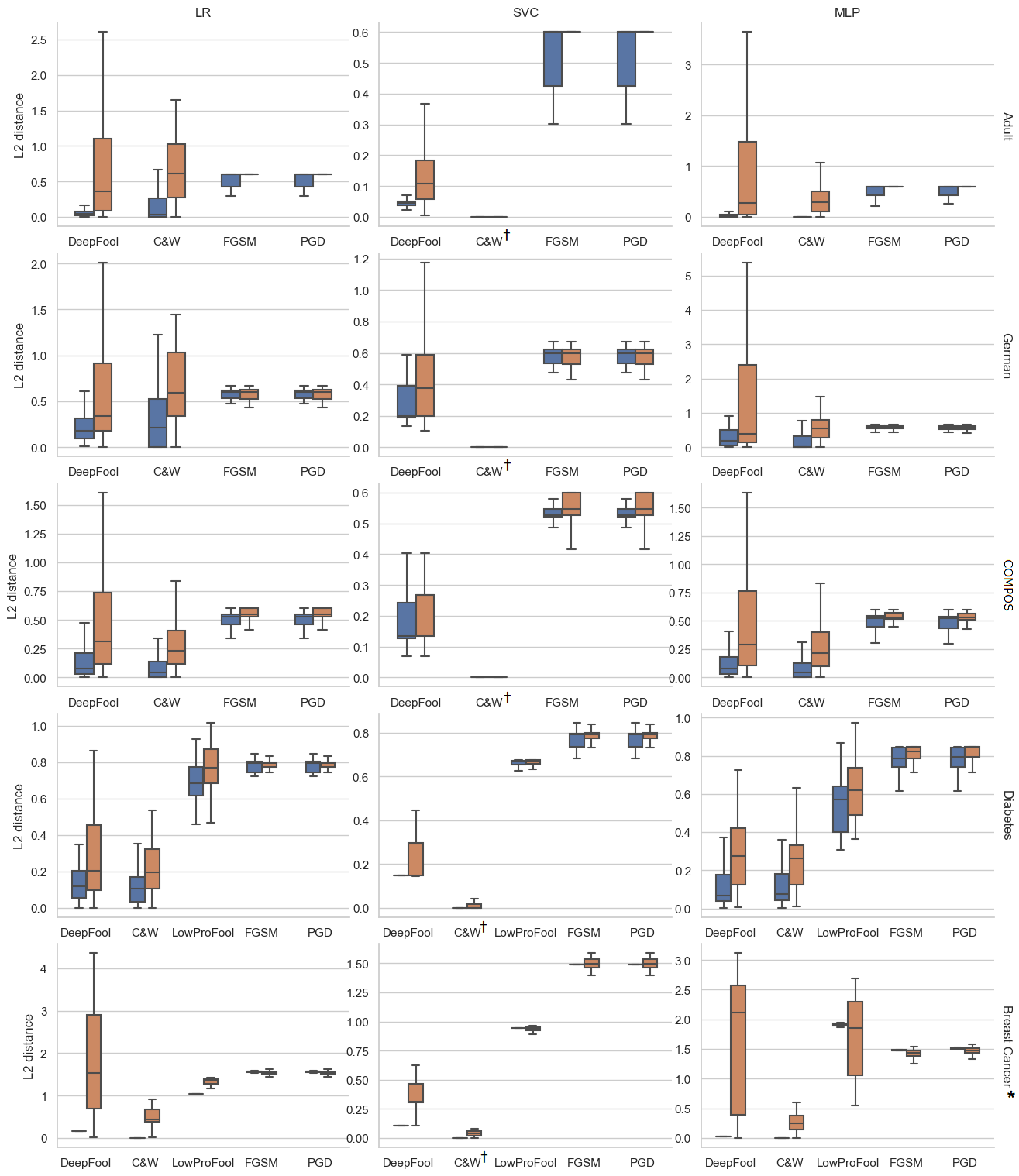}
    \caption{A series of box plots illustrating the distribution of \textbf{\emph{proximity \(\ell_2\)}} for both successful and unsuccessful attack examples. 
    The box plots are horizontally aligned for the same dataset and vertically aligned for the same predictive model. 
    In each box plot, the X-axis represents different adversarial attacks, while the Y-axis shows the distribution of \emph{proximity \(\ell_2\)}. 
    The {\color{MidnightBlue} blue} boxes represent {\color{MidnightBlue} unsuccessful attacks}, while the {\color{BurntOrange} orange} boxes indicate {\color{BurntOrange} successful attacks}. 
    Due to insufficient effectiveness, the combination of C\&W attack and LinearSVC is not considered and their results are marked with \(\dagger\).
    }
    \label{fig:trade-off:l2}
\end{figure*}

\begin{figure*}[]
    \centering
    \includegraphics[width=\textwidth]{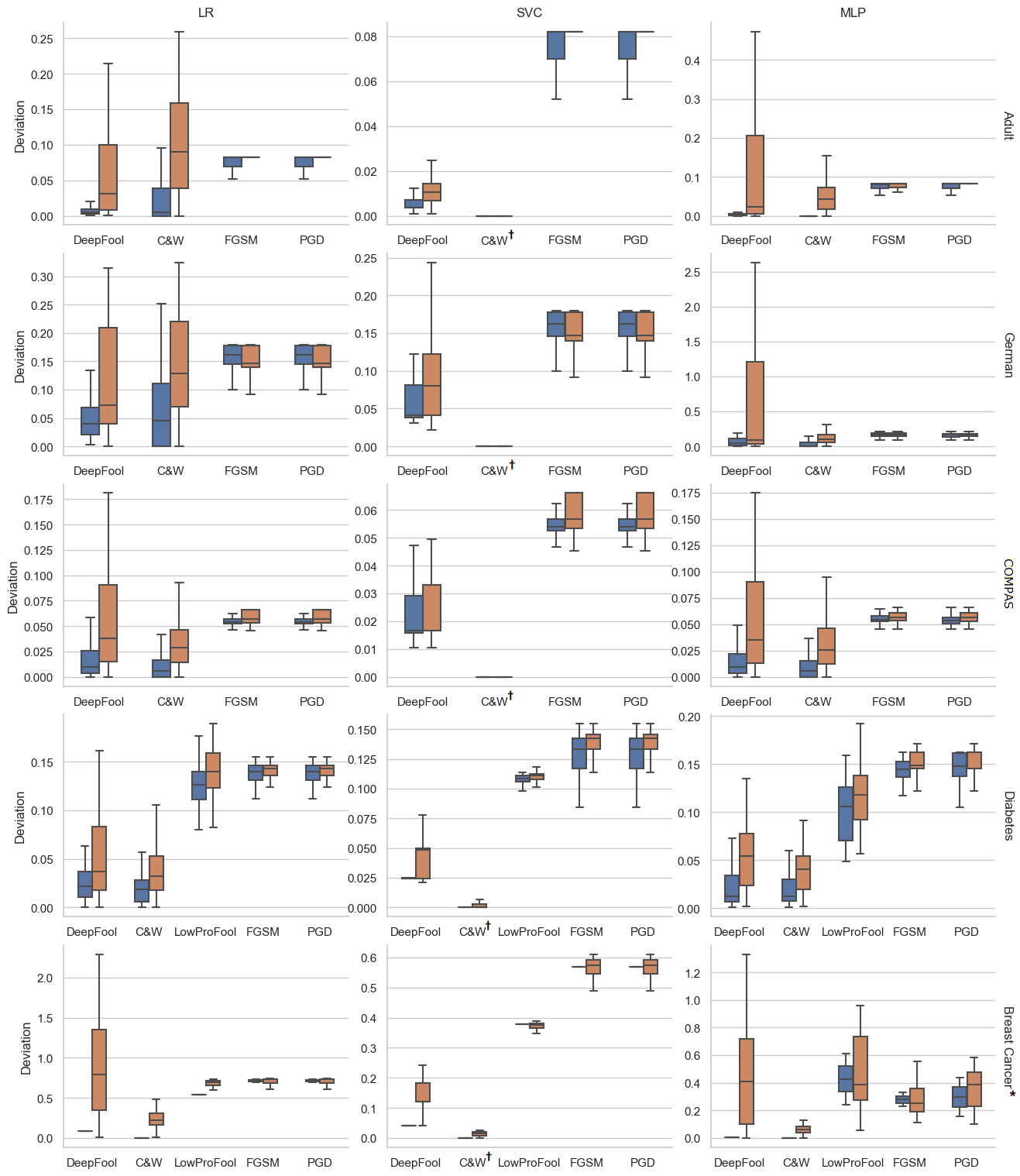}
    \caption{A series of box plots illustrating the distribution of \textbf{\textit{deviation}} for both successful and unsuccessful attack examples. The box plots are horizontally aligned for the same dataset and vertically aligned for the same predictive model. In each box plot, the X-axis represents different adversarial attacks, while the Y-axis shows the distribution of \textit{deviation}. The {\color{MidnightBlue} blue} boxes represent {\color{MidnightBlue} unsuccessful attacks}, while the {\color{BurntOrange} orange} boxes indicate {\color{BurntOrange} successful attacks}. Due to insufficient effectiveness, the combination of C\&W attack and LinearSVC is not considered and their results are marked with \(\dagger\). }
    \label{fig:trade-off:dev}
\end{figure*}

\begin{figure*}[]
    \centering
    \includegraphics[width=\textwidth]{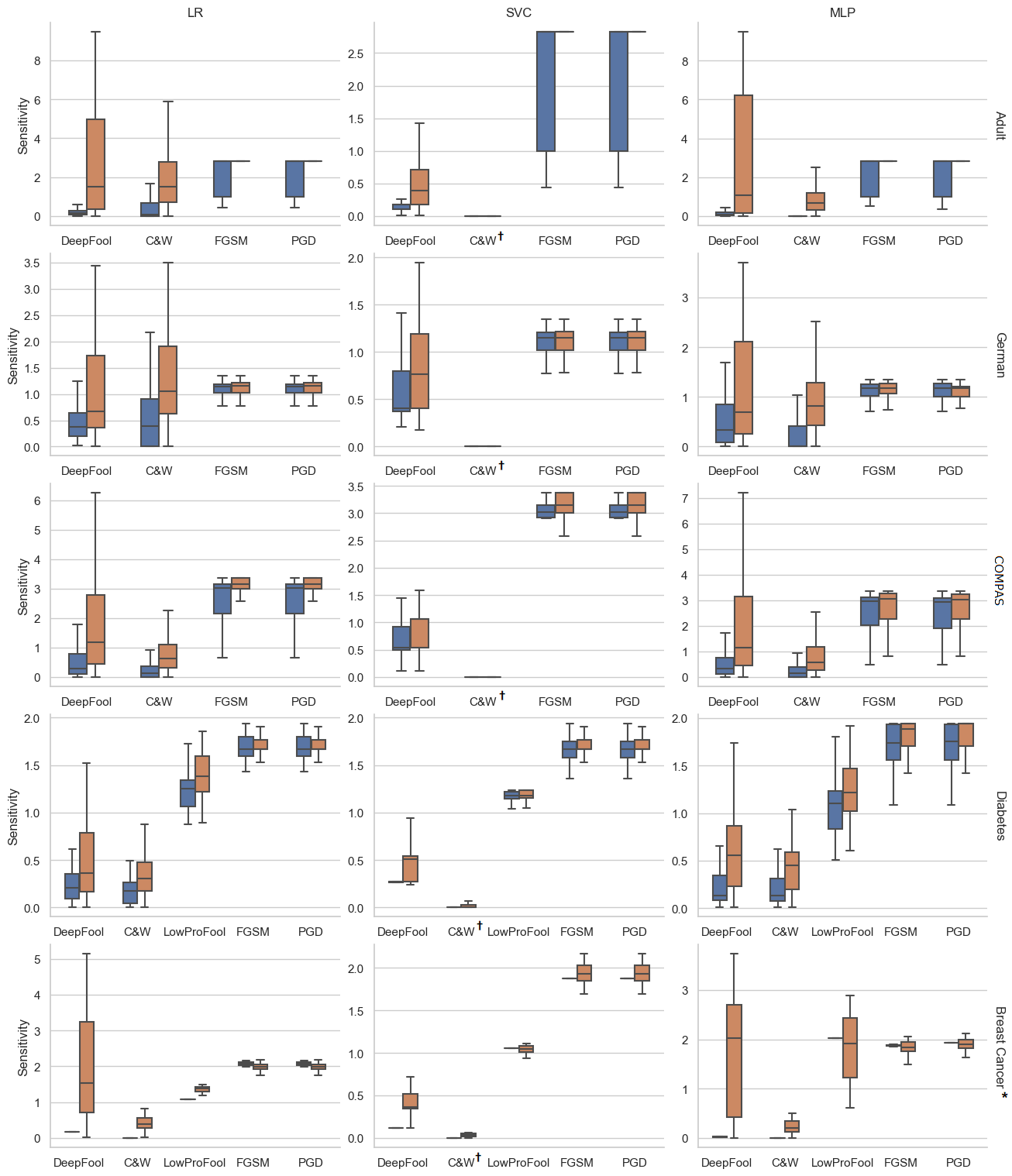}
    \caption{A series of box plots illustrating the distribution of \textbf{\textit{sensitivity}} for both successful and unsuccessful attack examples. The box plots are horizontally aligned for the same dataset and vertically aligned for the same predictive model. In each box plot, the X-axis represents different adversarial attacks, while the Y-axis shows the distribution of \textit{sensitivity}. The {\color{MidnightBlue} blue} boxes represent {\color{MidnightBlue} unsuccessful attacks}, while the {\color{BurntOrange} orange} boxes indicate {\color{BurntOrange} successful attacks}. Due to insufficient effectiveness, the combination of C\&W attack and LinearSVC is not considered and their results are marked with \(\dagger\).}
    \label{fig:trade-off:sens}
\end{figure*}

As depicted in Figure~\ref{fig:trade-off:l2}, the following insights regarding the relationship between proximity \(\ell_2\) and attack success can be drawn:

\begin{itemize}
    \item The high success rate on the Breast Cancer dataset leads to an imbalance between the number of successful and unsuccessful attack examples, making it unfair to compare their metric distributions. Therefore, this dataset is not consider in the following analysis and marked with \(\ast\) in the visualisation. 
    \item For the three unbounded attacks---DeepFool, C\&W and LowProFool, successful attacks tend to exhibit larger \emph{proximity \(\ell_2\)} values compared to unsuccessful attacks.
    \item For two bounded attacks---FGSM and PGD, their \emph{proximity \(\ell_2\)} values are constrained by the attack budget capping the maximum proximity. 
\end{itemize}

For proximity, the fixed attack budget in our experiments results in no difference in the \emph{proximity \(\ell_\infty\)} distribution between successful and unsuccessful examples for bounded attacks. The three unbounded attacks show similar results to those of \textit{proximity \(\ell_2\)}, that is, successful attacks yield larger \emph{proximity \(\ell_\infty\)} values compared to unsuccessful attacks.

Similar patterns can be observed from the results of \emph{deviation} (Figure~\ref{fig:trade-off:dev}) and \emph{sensitivity} (Figure~\ref{fig:trade-off:sens}), respectively.  
For \textit{deviation}, successful attacks tend to show larger values compared to unsuccessful attacks in the case of three unbounded attacks. This pattern also appears in two bounded attacks on the COMPAS and Diabetes datasets. Regarding \textit{sensitivity}, the values for successful attacks are generally higher than those for unsuccessful attacks, across all datasets for unbounded attacks and on the COMPAS and Diabetes datasets for bounded attacks.


Based on the above comparative analysis between attack success rate and three imperceptibility metrics, there is a trade-off between the effectiveness and imperceptibility of adversarial attacks on tabular data. 

\subsection{Challenges and Limitations}

\paragraph{Side effects of one-hot encoding}
Due to the use of \(\ell_p\)-based distance metrics in attack algorithms, encoding categorical variables when attacking tabular data is challenging. 
For example, ordinal encoding (or label encoding) can introduce an unintended order among categorical features. This unintended order often has an impact on the prediction results, as the attack algorithms might incorrectly infer that the numeric distance between categories reflects a true difference in the data. Given the challenges associated with ordinal encoding, one-hot encoding is our preferred encoding method for encoding categorical variables. 
However, one-hot encoding can increase the total number of features in a dataset as each unique category in a categorical feature is represented by a separate binary feature. This can lead to a high-dimensional feature space, which can result in computational challenges such as the curse of dimensionality~\citep{Borisov2021deep}. This necessitates a careful consideration of how distance metrics are applied to both numerical and categorical features to ensure effective and meaningful perturbation measurements in the pursuit of imperceptibility. Exploring alternative distance metrics, such as Gower's distance \citep{gower1971general} and other categorical feature similarity measures \citep{cost1993weighted,le2005association}, offers more options in encoding methods for categorical features in tabular data.

\paragraph{Focusing Solely on Proximity} 

Current adversarial attacks commonly focus on using the proximity of adversarial examples to the original inputs as the sole measure of imperceptibility. However, our evaluation highlights there are other essential properties for achieving imperceptible attacks but they are often overlooked by the evaluated attacks. Specifically, the design of current adversarial attacks does not adequately consider quantitative properties such as sparsity, sensitivity, and deviation, nor does it address qualitative aspects like immutability, feasibility, and feature interdependency. By integrating these properties into attack design, we can enhance the imperceptibility of adversarial attacks. For instance, our evaluation of sparsity indicates that all attack methods target the alteration of all numerical features, which are similar to how image-based attacks perturb all pixels. But as discussed in Section~\ref{sec:metric}, achieving imperceptibility in tabular data necessitates attacks to alter as few features as possible. Modifying fewer features is less likely to be noticeable to the human eye and thus harder to detect. Conversely, attacks that modify many features can significantly reduce imperceptibility, thereby increasing the risk of detection.

\paragraph{Assumption on uniform feature importance} 

Most current adversarial attacks operate under the assumption that all features contribute equally to predictive models, similar to the idea that each pixel in an image holds equal importance. However, in tabular datasets, the importance of different features is often not uniform due to their heterogeneous nature. Later in the research, there is a need for an extension into non-uniform adversarial attacks~\citep{erdemir2021adversarial,nandy2023non}. Addressing non-uniform adversarial challenges will contribute to a more comprehensive understanding of the robustness and generalisation capabilities of predictive models in practical applications.

\section{Conclusion}
\label{sec:conclusion}

In this paper, we have proposed seven key properties for evaluating the imperceptibility of adversarial attacks on tabular data, which are proximity, sparsity, deviation, sensitivity, immutability, feasibility, and feature interdependency. These properties can be used to comprehensively characterise imperceptible adversarial attacks on tabular data. 
Furthermore, we have conducted an empirical evaluation of imperceptibility for five adversarial attacks using the proposed metrics.  
The evaluation reveals that, except for proximity, these attacks often overlook the proposed imperceptibility properties in their algorithm design, and there is a trade-off between the effectiveness and imperceptibility of adversarial attacks. The insights gained from this evaluation offer valuable guidance for developing effective and imperceptible adversarial attacks on tabular data.

Our investigation of adversarial attacks on tabular data has provided valuable insights informing clear directions for the next stage of research. The use of one-hot encoding increases dataset dimensionality, highlighting the need for careful consideration of distance metrics for both numerical and categorical features. Additionally, the non-uniform importance of features in tabular data should be more thoroughly investigated to understand how varying feature significance impacts the design and effectiveness of adversarial attacks. Future research should expand to larger datasets, more sophisticated models, and a wider range of attack strategies.



%

\section*{Declaration of generative AI and AI-assisted technologies in the writing process}

During the preparation of this work the authors used ChatGPT for refining the language, enhancing clarity, and improving the overall coherence of the writing. After using this tool, the authors reviewed and edited the content as needed and take full responsibility for the content of the publication.



\section*{CRediT authorship contribution statement}

\textbf{Zhipeng He:} Conceptualization, Methodology, Software, Investigation, Writing -- Original Draft, Visualization. \textbf{Chun Ouyang:} Conceptualization, Methodology, Investigation, Writing -- Original Draft. \textbf{Laith Alzubaidi:} Supervision. \textbf{Alistair Barros:} Supervision. \textbf{Catarina Moreira:} Supervision, Investigation, Writing -- Original Draft.

\newpage

\appendix

\section{Qualitative Analysis Results}

\subsection{Immutability Results}\label{sec:appendix:immutable}

\begin{table}[h]
\centering
\caption{\textbf{Adult dataset}: Number of adversarial examples which perturb immutable feature \textit{race}, \textit{sex} and \textit{marital-status}. Total 6513 adversarial examples generated for each combination of attack algorithms and models.}
\label{tab:appendix:imm-1}
\begin{tabular}{@{}llrrr@{}}
\toprule
Model                      & Attack   & \multicolumn{1}{l}{\textit{race}} & \multicolumn{1}{l}{\textit{sex}} & \multicolumn{1}{l}{\textit{marital-status}} \\ \midrule
\multirow{4}{*}{LR}        & DeepFool & 17                         & 16                        & 472                                  \\
                           & C\&W     & 0                          & 0                         & 0                                    \\
                           & FGSM     & 0                          & 0                         & 0                                    \\
                           & PGD      & 0                          & 0                         & 0                                    \\ \midrule
\multirow{4}{*}{LinearSVC} & DeepFool & 0                          & 0                         & 0                                    \\
                           & C\&W     & 0                          & 0                         & 0                                    \\
                           & FGSM     & 0                          & 0                         & 0                                    \\
                           & PGD      & 0                          & 0                         & 0                                    \\ \midrule
\multirow{4}{*}{MLP}       & DeepFool & 648                        & 345                       & 1096                                 \\
                           & C\&W     & 0                          & 0                         & 0                                    \\
                           & FGSM     & 0                          & 0                         & 0                                    \\
                           & PGD      & 0                          & 0                         & 0                                    \\ \bottomrule
\end{tabular}
\end{table}
\begin{table}[h]
\centering
\caption{\textbf{Compas dataset}: Number of adversarial examples which perturb immutable feature \textit{race} and \textit{sex}. Total 1443 adversarial examples generated for each combination of attack algorithms and models.}
\label{tab:appendix:imm-2}
\begin{tabular}{@{}llrr@{}}
\toprule
Model                      & Attack   & \multicolumn{1}{l}{\textit{sex}} & \multicolumn{1}{l}{\textit{race}}  \\ \midrule
\multirow{4}{*}{LR}        & DeepFool & 0                          & 8    \\
                           & C\&W     & 0                          & 0    \\
                           & FGSM     & 0                          & 0    \\
                           & PGD      & 0                          & 0    \\ \midrule
\multirow{4}{*}{LinearSVC} & DeepFool & 0                          & 0    \\
                           & C\&W     & 0                          & 0    \\
                           & FGSM     & 0                          & 0    \\
                           & PGD      & 0                          & 0    \\ \midrule
\multirow{4}{*}{MLP}       & DeepFool & 36                         & 16   \\
                           & C\&W     & 0                          & 0    \\
                           & FGSM     & 0                          & 0    \\
                           & PGD      & 0                          & 0    \\ \bottomrule
\end{tabular}
\end{table}
\begin{table}[h]
\centering
\caption{\textbf{German dataset}: Number of adversarial examples which perturb immutable feature \textit{personal\_status\_sex} and \textit{foreign\_worker}. Total 200 adversarial examples generated for each combination of attack algorithms and models.}
\label{tab:appendix:imm-3}
\begin{tabular}{@{}llrr@{}}
\toprule
Model                      & Attack   & \multicolumn{1}{l}{\textit{personal\_status\_sex}} & \multicolumn{1}{l}{\textit{foreign\_worker}}  \\ \midrule
\multirow{4}{*}{LR}        & DeepFool & 0                          & 8    \\
                           & C\&W     & 0                          & 0    \\
                           & FGSM     & 0                          & 0    \\
                           & PGD      & 0                          & 0    \\ \midrule
\multirow{4}{*}{LinearSVC} & DeepFool & 0                          & 0    \\
                           & C\&W     & 0                          & 0    \\
                           & FGSM     & 0                          & 0    \\
                           & PGD      & 0                          & 0    \\ \midrule
\multirow{4}{*}{MLP}       & DeepFool & 36                         & 16   \\
                           & C\&W     & 0                          & 0    \\
                           & FGSM     & 0                          & 0    \\
                           & PGD      & 0                          & 0    \\ \bottomrule
\end{tabular}
\end{table}
\newpage

\subsection{Feasible Feature Range for Feasibility}\label{sec:appendix:feasible}

\begin{table}[h!]
\centering
\caption{Description for features in Diabetes dataset and their feasible value ranges from medical domain knowledge.}
\label{tab:diabetes-knowledge}
\begin{tabular}{@{}lll@{}}
\toprule
\textbf{Feature}                         & \textbf{Description}                                              & \textbf{Range}                                                  \\ \midrule
\multirow{3}{*}{Glucose}        & \multirow{3}{*}{2-hours glucose tolerance test results} & $<$140 mg/dL is considered normal               \\
                                &                                                         & 140--199 mg/dL indicates impaired glucose tolerance     \\
                                &                                                         & $\geq$200 mg/dL indicates diabetes                          \\ \midrule
\multirow{2}{*}{Blood Pressure} & \multirow{2}{*}{Diastolic blood pressure}               & $<$80 mm HG is considered normal                \\
                                &                                                         & $\geq$80 mm HG indicates high blood pressure \\ \midrule
SkinThickness                   & Triceps skin fold thickness                             & 23.6$\pm$7.5mm                                         \\ \midrule
Insulin                         & 2-Hour serum insulin                                    & 16-166 mIU/L                                           \\ \midrule
\multirow{6}{*}{BMI}            & \multirow{6}{*}{Body Mass Index}                         & Less than 18.5 (Underweight)                           \\
                                &                                                         & 18.5 to 24.9 (Healthy)                                 \\
                                &                                                         & 25 to 29.9 (Overweight)                                \\
                                &                                                         & 30 to 34.9 Obese class I                               \\
                                &                                                         & 35 to 39.9 Obese class II                              \\
                                &                                                         & 40 or more Obese class III: severe obesity             \\ \bottomrule
\end{tabular}
\end{table}

\newpage

\bibliographystyle{elsarticle-harv} 
\bibliography{cas-refs}





\end{document}